\documentclass[lettersize,journal]{IEEEtran}
\usepackage{amsmath,amsfonts}
\usepackage{algorithmic}
\usepackage{algorithm}
\usepackage{array}
\usepackage[caption=false,font=normalsize,labelfont=sf,textfont=sf]{subfig}
\usepackage{textcomp}
\usepackage{stfloats}
\usepackage{url}
\usepackage{verbatim}
\usepackage{graphicx}
\usepackage{cite}
\usepackage{booktabs}       
\usepackage{xcolor}         
\usepackage{colortbl}       
\usepackage{multirow}       
\usepackage{graphicx}       
\usepackage{amssymb}

\begin{document}

\title{Autoregression-Free Neural Operators for Time-Dependent PDEs}
\author{Jiaquan Zhang, Caiyan Qin, Haoyu Bian, Libin Cai, Yi Lu, Chaoning Zhang,~\IEEEmembership{Senior Member, IEEE}, Wei Dong, Yuanfang Guo,~\IEEEmembership{Senior Member, IEEE}  Yang Yang,~\IEEEmembership{Senior Member, IEEE} and Heng Tao Shen,~\IEEEmembership{Fellow, IEEE}

\thanks{Jiaquan Zhang, Haoyu Bian, Libin Cai, Chaoning Zhang and Yang Yang are with the School of Computer Science and Engineering, University of Electronic Science and Technology of China, Chengdu, Sichuan 611731, China (email: jiaquanzhang2005@gmail.com; haoyubian04@gmail.com; 2024091202004@std.uestc.edu.cn; chaoningzhang@uestc.edu.cn; yang.yang@uestc.edu.cn).

Caiyan Qin is with the School of Robotics and Advanced Manufacture, Harbin Institute of Technology, Shenzhen, Guangdong 518055, China (email: qincaiyan@hit.edu.cn). 

Yi Lu is with the School of Mathematical Sciences, Capital Normal University, Beijing 100048, China (email: ylulu0610@gmail.com).

Wei Dong is with the College of Information and Control Engineering,
Xi’an University of Architecture and Technology, Xi’an, Shaanxi 710064, China (email: dongwei156@outlook.com).

Yuanfang Guo is with the Laboratory of Intelligent Recognition and Image Processing, School of Computer Science and Engineering, Beihang University, Beijing 100191, China (e-mail: andyguo@buaa.edu.cn).

Heng Tao Shen is with the School of Computer Science and Technology, Tongji University, Shanghai 201804, China (e-mail: shenhengtao@tongji.edu.cn).
}

\thanks{Manuscript received April 19, 2021; revised August 16, 2021.}}

\markboth{Journal of \LaTeX\ Class Files,~Vol.~14, No.~8, August~2021}%
{Shell \MakeLowercase{\textit{et al.}}: A Sample Article Using IEEEtran.cls for IEEE Journals}


\maketitle

\begin{abstract}
Neural operators learn mappings from function-dependent inputs to solutions, providing an effective framework for solving partial differential equations (PDEs). 
For time-dependent PDEs, existing methods typically perform long-horizon prediction through autoregressive rollout directly in high-dimensional physical field spaces, where each predicted state is recursively fed back as the input for the next step. Although effective for short-term prediction, this autoregressive rollout and the lack of continuous-time modeling lead to progressive error accumulation over long-horizon rollouts. In this work, we propose Autoregression-Free Neural Operators (AFNO), which map the time evolution of PDEs into a latent space and model continuous-time vector fields within it. AFNO uses flow matching to learn the latent vector field, thereby enabling continuous evolution over extended horizons, avoiding autoregressive rollout and capturing dynamics under varying parameter configurations through explicit conditioning on physical parameters.
Theoretical analysis and extensive experiments on six PDEs demonstrate that AFNO improves long-horizon prediction stability and consistently reduces rollout errors compared with the baselines. 


\end{abstract}


\section{Introduction}
\label{sec:1}
\IEEEPARstart{T}{ime-dependent} partial differential equations (PDEs) are fundamental in scientific computing and physical simulation, with applications in fluid mechanics, weather prediction, and material modeling \cite{sharma2023review,mengaldo2019current,patawari2025traditional}. Traditional numerical solvers rely on discretized time stepping and parameter-specific solves, resulting in high computational cost and limited scalability \cite{tadmor2012review}. In recent years, deep learning methods for solving PDEs have achieved significant progress \cite{huang2025partial}, with neural operators learning mappings between function spaces and providing a new framework for equation modeling \cite{azizzadenesheli2024neural,kovachki2023neural}. For time-dependent PDEs, the central objective is to learn the temporal evolution of physical fields, such as transport and diffusion in fluids \cite{zhou2023physics}, wave propagation \cite{diab2025temporal}, and phase evolution in materials \cite{oommen2024rethinking,meng2025physics}. 
Such a task requires approximating solution operators that propagate physical states across time, rather than merely predicting a static solution field. 
Methods such as Fourier Neural Operator (FNO) \cite{li2020fourier} address temporal evolution modeling by learning time-advancement operators in high-dimensional physical fields, enabling data-driven multi-step prediction. 
Subsequent studies incorporate self-supervised and multi-resolution training, together with low-rank structures and physical constraints \cite{song2024seismic,liu2023domain,you2025mscalefno}. 
These advances improve generalization across complex geometries, multiple physical fields, and high-dimensional settings, thereby broadening the applicability of neural operators to fluid simulation, weather forecasting, seismic modeling, and material design.

\begin{figure}[t]
  \centering
  \includegraphics[width=\linewidth]{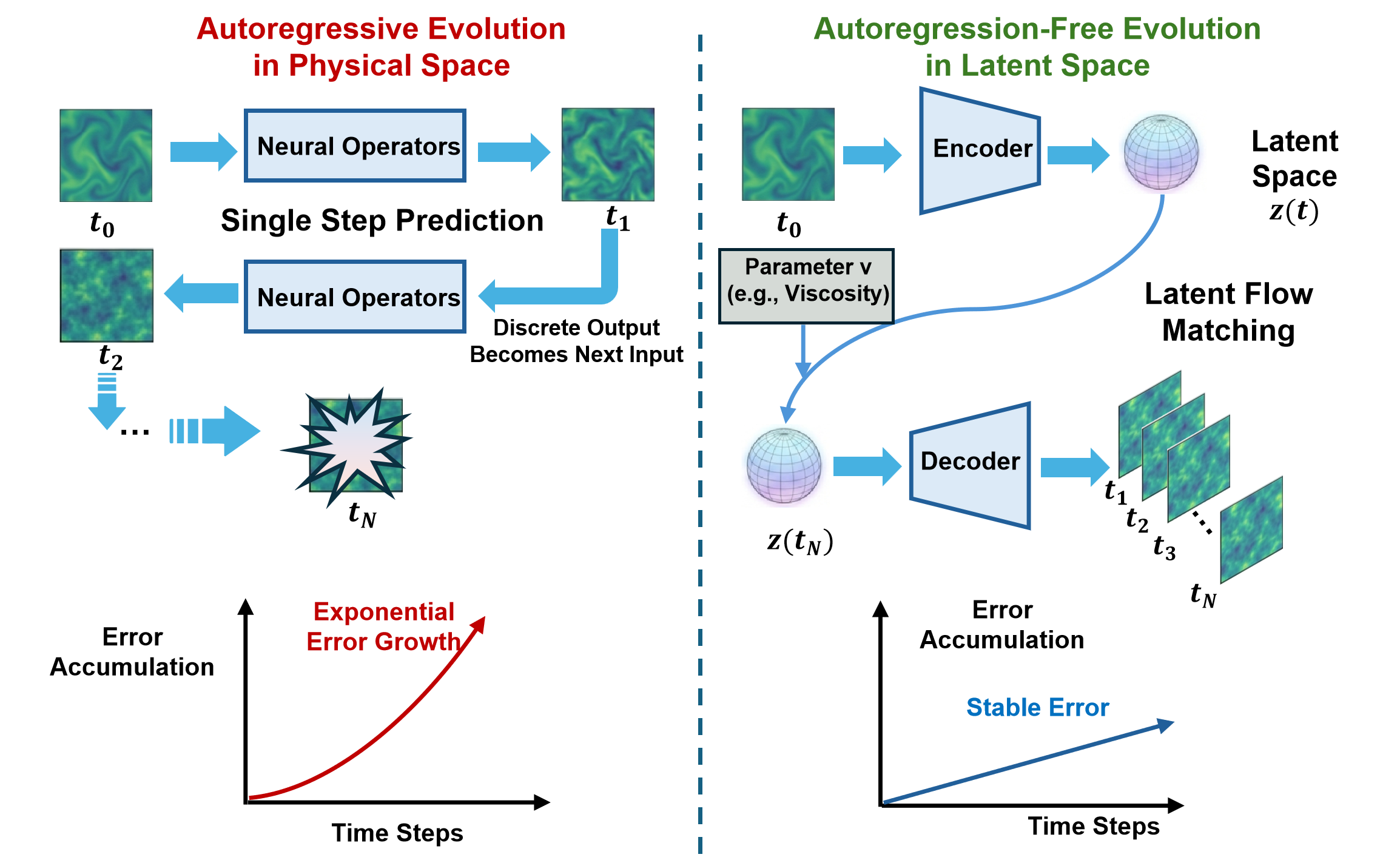}
    \caption{Comparison of autoregressive rollout and autoregression-free rollout for time-dependent PDEs. (Left) Existing methods perform autoregressive rollout in high-dimensional physical space, where predictions are recursively fed back as inputs, leading to amplified error accumulation over long horizons. (Right) Our method encodes physical fields into a parameter-conditioned latent space, with continuous latent space rollout via flow-matching dynamics yielding stable error propagation.}
  \label{fig:1}
\end{figure}

However, most existing methods perform long-horizon prediction in an autoregressive manner, where the predicted state at each step is recursively fed back as the input for the next step \cite{tanyu2023deep}, as illustrated in Fig.~\ref{fig:1}. Although effective for short-term forecasting, this strategy causes the mismatch between prediction and ground truth to be propagated through subsequent rollout steps during long-horizon extrapolation, resulting in error accumulation and prediction instability. This effect is particularly severe in high-dimensional physical fields with complex spatial coupling and multi-scale structures, where local errors spread spatially and compound across scales. For example, in turbulent flow, the coupling of velocity and pressure across multiple scales leads to rapid error accumulation \cite{wang2020towards}. Moreover, discrete time stepping for inherently continuous-time PDE evolution obscures the underlying system structure and limits adaptability to varying time steps and multiple temporal scales \cite{sun2020neupde}. Beyond this issue, physical parameters such as viscosity, diffusion strength, and time step are typically treated implicitly. This further restricts generalization across different parameter regimes, e.g., varying Reynolds numbers in fluid simulations \cite{gupta2022three}.

These limitations motivate us to ask whether long-horizon PDE prediction can be made autoregression-free. 
Recent advances in generative modeling, such as diffusion \cite{croitoru2023diffusion}, latent diffusion models (LDMs) \cite{rombach2022high}, and flow-based models \cite{zhao2024flowturbo}, describe temporal variation as continuous trajectories evolving within a structured latent space \cite{elasri2022image}. 
In particular, LDMs encode high-dimensional data into a compact latent space, thereby reformulating high-dimensional data modeling over structured latent representations rather than raw observations.
\cite{croitoru2023diffusion}. 
This motivates us to transfer PDE temporal evolution from high-dimensional physical fields to a latent dynamical space, where essential dynamics can be modeled more compactly and stably.
Under this formulation, long-horizon prediction is performed by numerically integrating the latent vector field rather than by recursively feeding predicted physical states back into a neural operator.
By decoupling intrinsic dynamics from high-dimensional physical observations, the latent space facilitates the extraction of coherent temporal features.
It further provides a compact manifold for consistent parameterization of temporal evolution.
Flow matching supervises the local vector field through finite-difference derivative alignment, thereby improving multi-step rollout stability.
The conceptual shift from autoregressive physical space rollout to our paradigm is illustrated in Fig.~\ref{fig:1}.

Motivated by this insight, we propose Autoregression-Free Neural Operators (AFNO), a framework for modeling PDE temporal evolution in a restructured latent space. Specifically, AFNO first employs an FNO-based autoencoder to reorganize high-dimensional spatial fields, mapping physical fields to latent representations that preserve multi-scale structures. Next, in the latent space, we perform multi-step extrapolation on the discrete observed data via explicit numerical integration. To this end, a latent flow matching module is designed to supervise the latent vector field using finite-difference temporal derivatives extracted from latent trajectories, thereby enabling stable continuous-time evolution through numerical integration. AFNO also explicitly conditions the latent vector field on physical parameters and time steps through learned embeddings, enabling adaptive dynamics across parameter configurations within the same PDE family. Finally, AFNO decodes latent representations back to the physical space by exploiting inverse fast Fourier transform and the symmetry of the FNO architecture. The entire framework is trained end-to-end, jointly optimizing latent encoding, latent flow modeling, and decoding. Extensive experiments on six PDEBench \cite{takamoto2022pdebench} benchmarks demonstrate that AFNO outperforms baseline models in terms of accuracy and long-horizon stability. In particular, AFNO exhibits more gradual error growth and stronger prediction stability during long-horizon rollouts.
At 100 rollout steps, it reduces the final mean squared error (MSE) and relative $L_2$ error by 50--85\% compared with the best-performing baseline.
We further compare AFNO with recent methods designed for long-horizon rollout stability, and AFNO reduces the final-step MSE and relative $L_2$ error by over 90\% against the best-performing baseline. Moreover, theoretical analysis indicates that AFNO alleviates the recursive error amplification of autoregressive physical space rollout by replacing repeated operator application with continuous latent space evolution. Our contributions are as follows:
\begin{itemize}
    \item We introduce an autoregression-free operator learning paradigm for time-dependent PDEs, reformulating long-horizon prediction from recursive physical space rollout to latent dynamics modeling.
    \item Based on this paradigm, we develop AFNO, a parameter-conditioned autoregression-free neural operator that combines structured latent representations with flow matching for stable latent space evolution.
    \item We provide both theoretical analysis and extensive empirical evaluation, validating the improved long-horizon prediction accuracy and stability of AFNO.
\end{itemize}

\section{Related Work}
\label{sec:2}
We review related work on neural operators and latent representation learning, which are closely connected to long-horizon prediction and latent space dynamics for time-dependent PDEs.

\subsection{Neural Operators}
\label{sec:2.1}
Neural operators formulate PDE solution prediction as a generalizable mapping across initial conditions, boundary conditions, and parameter regimes. Representative models approximate solution operators for general PDE-solving tasks. DeepONet~\cite{lu2019deeponet} parameterizes operators via a branch--trunk architecture for functional inputs. FNO~\cite{li2020fourier} further introduces Fourier-domain operator parameterization, enabling efficient modeling of global interactions and zero-shot super-resolution in turbulent flow simulation.
From a functional approximation perspective, the approach of Li et al.~\cite{li2020neural} formalizes infinite-dimensional mappings via combinations of nonlinear activations and integral operators, computing kernel integrals through message passing on graphs. Subsequent extensions improve neural operators for more complex PDE settings. Geo-FNO \cite{li2023fourier} incorporates coordinate transformations to handle irregular boundaries and complex geometries. Physics-Informed Neural Operator (PINO) \cite{li2024physics} combines neural operators with Physics-Informed Neural Networks (PINNs) \cite{cuomo2022scientific} by introducing PDE residuals into the loss, enforcing physical consistency while learning operator mappings. 


For time-dependent PDEs, neural operators are often used as time-advancement models that predict future physical states from current or historical states. 
Convolutional Neural Operator (CNO)~\cite{raonic2023convolutional} employs anti-aliased convolutional operators for continuous signals, improving accuracy in complex spatiotemporal fields. 
GNOT~\cite{hao2023gnot} leverages Transformer~\cite{vaswani2017attention} architectures to handle heterogeneous inputs, dynamic boundary conditions, and multi-physics coupling. 
Mamba Neural Operator (MNO)~\cite{cheng2025mamba} exploits the linear-complexity sequence modeling capability of Mamba~\cite{gu2024mamba} to efficiently capture long-horizon PDE dynamics. More recently, long-horizon rollout stability has received increasing attention in neural operator learning. Recurrent Neural Operators (RNO) \cite{ye2025recurrent} improve long-horizon prediction by aligning training with recursive inference, while Spectral Generator Neural Operators (SGNO) \cite{li2026sgno} enhance rollout stability via constrained spectral evolution and controlled nonlinear updates. However, RNO and SGNO remain largely within autoregressive rollout frameworks. As a result, prediction errors may continue to accumulate in the high-dimensional physical field space over long horizons.

\subsection{Latent Representation Learning}
\label{sec:2.2}
Latent space learning constitutes a fundamental paradigm in machine learning \cite{dao2023flow}, underpinning tasks such as image generation \cite{duan2024weditgan} and representation learning \cite{gelada2019deepmdp}. The core idea of latent space learning projects high-dimensional observations or data into a low-dimensional, structured latent space, forming a representation paradigm that enhances generalization, robustness, and interpretability. In image generation, Variational Autoencoders (VAE) \cite{kingma2013auto} encode images into continuous latent spaces for controllable generation and reconstruction; Generative Adversarial Networks (GAN) \cite{goodfellow2020generative} and flow-based models such as RealNVP \cite{dinh2016density} and Glow \cite{kingma2018glow} learn latent distributions to generate high-quality images; and diffusion models such as DDPM \cite{ho2020denoising} progressively model data distributions in latent or feature spaces to enable continuous sampling. LDMs \cite{rombach2022high} further shift the generative process from the high-dimensional data space to a compact latent space, demonstrating the effectiveness of structured latent representations for modeling complex data distributions. In natural language processing, models like BERT \cite{devlin2019bert} learn latent embeddings of sentences or documents to support text generation, semantic retrieval, and dialogue systems. In speech and audio generation, models such as VQ-VAE \cite{van2017neural} and WaveNet \cite{van2016wavenet} leverage latent representations to capture acoustic features, enabling high-fidelity synthesis and style transfer. 

Recently, latent space methods have also advanced physical simulation and scientific computing. For instance, Latent Neural Operator (LNO) 
\cite{wang2024latent} learns physical mappings in a latent space, bridging geometry and latent representations via Physics-Cross-Attention to accelerate operator inference. While LNO uses latent representations mainly to improve the efficiency of operator approximation and inference for physical mappings, rather than to model temporal dynamics or long-horizon error propagation.
LDMs, on the other hand, mainly exploit latent spaces for generative modeling.
In contrast, AFNO constructs a latent dynamical space for time-dependent PDEs to improve the stability of long-horizon extrapolation.

\section{Theoretical Analysis and Motivation}
\label{sec:3}

This section analyzes error propagation in autoregressive neural operators and motivates the proposed formulation.
Detailed theoretical derivations are provided in the Supplementary Material.

\subsection{Error Amplification in Autoregressive Rollout}

Let $\mathcal{U}$ denote the function space of admissible physical states. The temporal evolution of a well-posed time-dependent PDE induces a solution operator family $\{\Phi_t\}_{t\geq 0}$ on $\mathcal{U}$, where $\Phi_t:\mathcal{U}\rightarrow\mathcal{U}$ maps an initial state to its state after time $t$. This family satisfies the semigroup property
\begin{equation}
\Phi_0=\mathrm{Id}, \qquad 
\Phi_{s+t}=\Phi_s\circ\Phi_t,\quad \forall s,t\geq 0.
\end{equation}
Thus, for a uniform time step $\Delta t$, the exact long-horizon evolution over $n$ steps is
\begin{equation}
\Phi_{n\Delta t}=(\Phi_{\Delta t})^n.
\end{equation}

Existing neural operators typically approximate the short-time flow map $\Phi_{\Delta t}$ by a learned one-step propagator $F:\mathcal{U}\rightarrow\mathcal{U}$. During inference, long-horizon prediction is performed autoregressively:
\begin{equation}
\hat u_{t+\Delta t}=F(\hat u_t),
\end{equation}
whereas the exact dynamics satisfy
\begin{equation}
u_{t+\Delta t}=\Phi_{\Delta t}(u_t).
\end{equation}
Let $e(t)=\hat u_t-u_t$ denote the rollout error.

\noindent\textbf{Assumption 1.}
The learned one-step propagator $F$ is Lipschitz continuous on $\mathcal{U}$; that is, there exists a constant $L_F>0$ such that
\begin{equation}
\|F(u)-F(v)\|\leq L_F\|u-v\|,\qquad \forall u,v\in\mathcal{U}.
\end{equation}

\noindent\textbf{Proposition 1.}
Under Assumption 1, let 
\begin{equation}
\epsilon_t=\|F(u_t)-\Phi_{\Delta t}(u_t)\|
\end{equation}
denote the local one-step approximation error. After $n$ autoregressive rollout steps, the accumulated error satisfies
\begin{equation}
\|e(t+n\Delta t)\|
\leq
L_F^n\|e(t)\|
+
\sum_{i=0}^{n-1}
L_F^i
\epsilon_{t+(n-1-i)\Delta t}.
\end{equation}

\noindent\textbf{Proof sketch 1.}
For one rollout step, adding and subtracting $F(u_t)$ gives
\begin{equation}
\begin{aligned}
e(t+\Delta t)
&=
F(\hat u_t)-\Phi_{\Delta t}(u_t) \\
&=
\underbrace{F(\hat u_t)-F(u_t)}_{\text{input perturbation}}
+
\underbrace{F(u_t)-\Phi_{\Delta t}(u_t)}_{\text{local approximation error}} .
\end{aligned}
\end{equation}
Taking norms and applying Assumption 1 yields
\begin{equation}
\|e(t+\Delta t)\|
\leq
L_F\|e(t)\|+\epsilon_t .
\end{equation}
The multi-step bound follows by recursively applying this inequality.

\noindent\textbf{Remark 1.}
Proposition 1 shows that autoregressive rollout amplifies both the initial prediction error and the accumulated local approximation errors through repeated physical space propagation. In particular, the homogeneous term scales as $L_F^n$, which may grow rapidly when $L_F>1$. For encoder--decoder rollouts, physical space feedback can further propagate the reconstruction residual $r(u)=\mathrm{Dec}(\mathrm{Enc}(u))-u$ once decoded predictions are reused as inputs. This compounding-error effect is a key limitation of autoregressive long-horizon prediction~\cite{bengio2015scheduled,ross2011reduction,ye2025recurrent}.

\subsection{Latent Space Error Propagation of AFNO}

AFNO avoids recursive feedback in the physical field space by evolving trajectories in latent space after the initial encoding and using the decoder only for the readout of physical-state, consistent with continuous latent dynamics modeling~\cite{chen2018neural,rubanova2019latent}.

\noindent\textbf{Proposition 2.}
Let $z_t=\mathrm{Enc}(u_t)$ denote the latent representation of the physical state, and let the latent evolution of AFNO be modeled by
\begin{equation}
\hat z_{t+\Delta t}
=
\hat z_t+\Delta t\,v_\theta(\hat z_t,b),
\end{equation}
where $b=b(\nu,\Delta t)$ is the parameter embedding.
Assume that $v_\theta(\cdot,b)$ is Lipschitz continuous with constant $L_v$, and that $\mathrm{Dec}$ is Lipschitz continuous with constant $L_{\mathrm{dec}}$.
Let $e_z(t)=\hat z_t-z_t$ be the latent rollout error, and define the local latent consistency error as
\begin{equation}
\epsilon_t^z
=
\left\|
z_t+\Delta t\,v_\theta(z_t,b)-\Psi_{\Delta t}(z_t)
\right\|,
\end{equation}
where $\Psi_{\Delta t}$ denotes the exact latent one-step flow.
Then
\begin{equation}
\|e_z(t+\Delta t)\|
\leq
(1+\Delta t L_v)\|e_z(t)\|+\epsilon_t^z .
\end{equation}

After $n$ latent rollout steps, the accumulated latent error satisfies
\begin{equation}
\begin{aligned}
\|e_z(t+n\Delta t)\|
&\leq
(1+\Delta t L_v)^n\|e_z(t)\| \\
&\quad+
\sum_{i=0}^{n-1}
(1+\Delta t L_v)^i
\epsilon^z_{t+(n-1-i)\Delta t}.
\end{aligned}
\end{equation}

The corresponding physical space error is bounded by
\begin{equation}
\begin{aligned}
\|\hat u_{t+n\Delta t}-u_{t+n\Delta t}\|
&\leq
L_{\mathrm{dec}}\|e_z(t+n\Delta t)\| \\
&\quad+
\epsilon_{\mathrm{rec}}(t+n\Delta t).
\end{aligned}
\end{equation}
where $\epsilon_{\mathrm{rec}}(t)=\|\mathrm{Dec}(\mathrm{Enc}(u_t))-u_t\|$ denotes the reconstruction error.

\noindent\textbf{Proof sketch 2.}
By adding and subtracting $z_t+\Delta t\,v_\theta(z_t,b)$, we obtain
\begin{equation}
\begin{aligned}
e_z(t+\Delta t)
=&
\hat z_t-z_t
+
\Delta t\bigl(v_\theta(\hat z_t,b)-v_\theta(z_t,b)\bigr) \\
&+
z_t+\Delta t\,v_\theta(z_t,b)-\Psi_{\Delta t}(z_t).
\end{aligned}
\end{equation}

Taking norms and applying the Lipschitz continuity of $v_\theta$ gives
\begin{equation}
\|e_z(t+\Delta t)\|
\leq
(1+\Delta tL_v)\|e_z(t)\|+\epsilon_t^z .
\end{equation}

The multi-step bound follows by recursively applying this inequality. The physical space bound follows from the Lipschitz continuity of $\mathrm{Dec}$ and the reconstruction residual.

\noindent\textbf{Remark 2.}
Proposition 2 shows that AFNO changes the error propagation structure of long-horizon prediction. In the physical space rollout, the complete one-step error is repeatedly fed back into subsequent inputs, leading to the amplification factor $L_F^n$ in Proposition 1. In contrast, AFNO performs intermediate rollout in latent space, where the amplification factor becomes $(1+\Delta tL_v)^n$. When $\Delta tL_v\ll 1$, this term satisfies
\begin{equation}
(1+\Delta tL_v)^n
\approx
\exp(L_vT),
\qquad T=n\Delta t,
\end{equation}
which indicates that the homogeneous latent error growth is governed by the physical prediction horizon and the regularity of the learned vector field.

\begin{figure}[t]
  \centering
  \includegraphics[width=\linewidth]{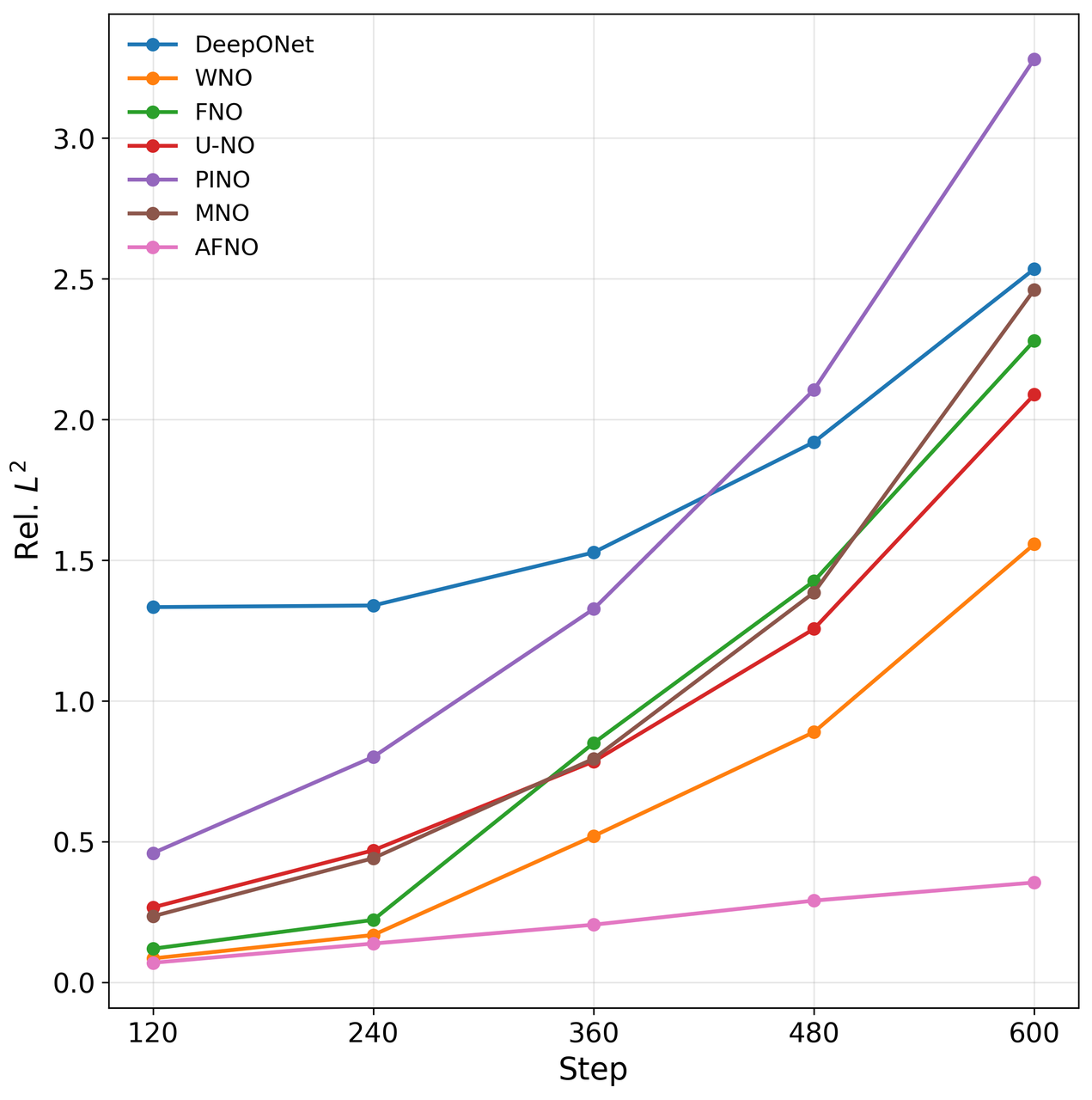}
  \caption{Motivating case study of long-horizon rollout error on the 1D-Burgers equation. As the prediction horizon increases, autoregressive baselines exhibit progressively steeper relative $L_2$ error growth, whereas AFNO shows a more gradual trend.}
  \label{fig:case}
\end{figure}

\subsection{Empirical Motivation}
The empirical behavior in Fig.~\ref{fig:case} corroborates the error propagation analysis above. As the rollout horizon increases from 120 to 600 steps, the relative $L_2$ error rises for all compared methods, while the growth profiles differ markedly. Autoregressive baselines exhibit increasingly steep error curves at longer horizons, which is characteristic of approximately exponential error accumulation induced by recursive amplification of errors under repeated one-step propagation in the physical space. In contrast, AFNO maintains the lowest error throughout the evaluation range and exhibits a slow, stable, and approximately linear growth trend.

\section{Method}
\label{sec:4}
Autoregression-Free Neural Operators (AFNO) models the evolution of time-dependent PDEs in a latent space, where a parameter-conditioned vector field is learned from discrete observations via flow matching.

\begin{figure*}[t]
  \centering
  \includegraphics[width=\linewidth]{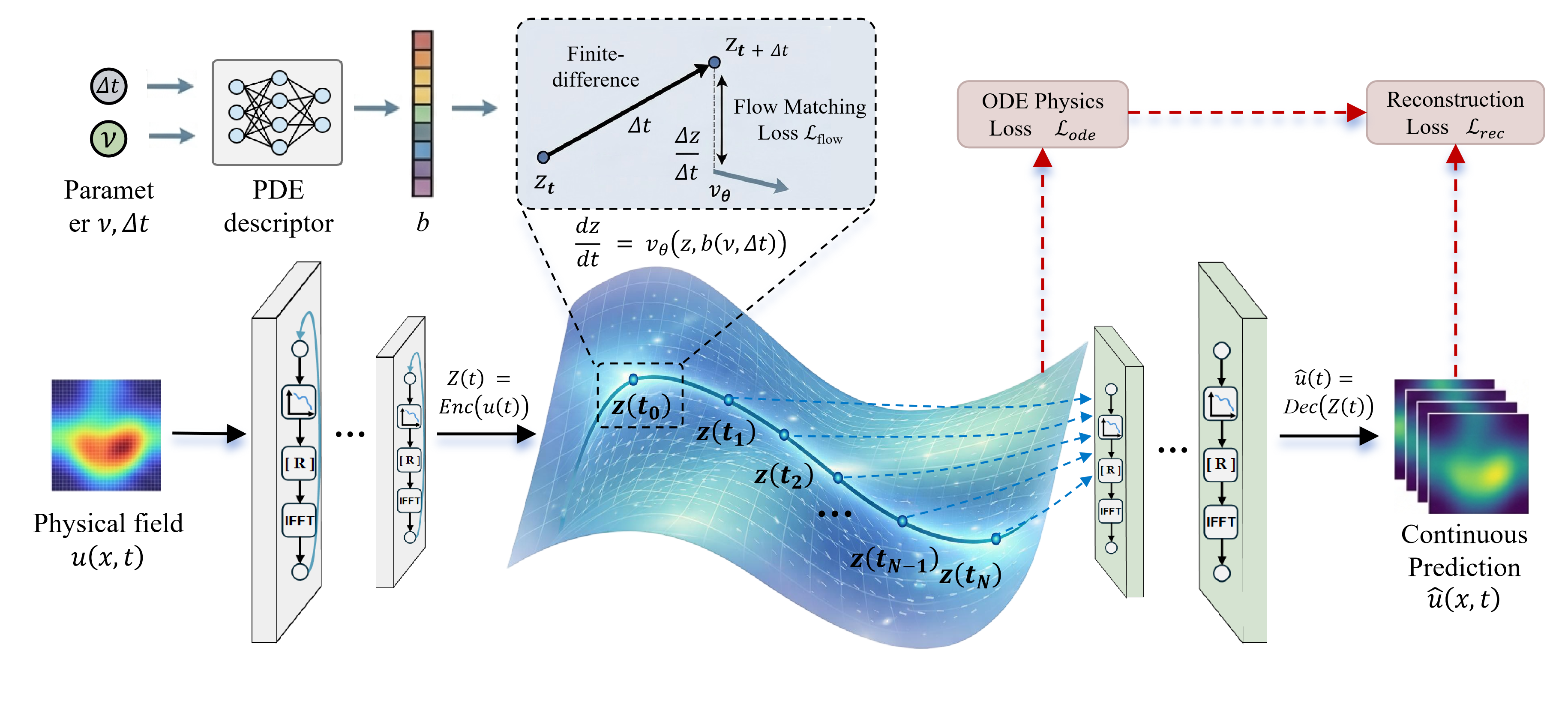}
   \caption{Overview of the Autoregression-Free Neural Operators (AFNO). The input physical field is encoded into a compact latent representation using an FNO-based encoder. Physical parameters are embedded and injected into the latent space, where a vector field governs temporal evolution through explicit ODE integration. The updated latent state is decoded back to the predicted physical field.}
  \label{fig:main}
\end{figure*}

\subsection{AFNO Architecture Overview}
\label{sec:4.1}
AFNO consists of three main components: a latent representation module, a latent flow learning module, and a parameter conditioning mechanism. As illustrated in Fig.~\ref{fig:main}, given an input physical field $u(\mathbf{x},t)$ defined on a spatial domain $\mathbf{x}\in\Omega\subset\mathbb{R}^d$, AFNO first employs an FNO-based autoencoder to map the high-dimensional physical field $u_t := u(\mathbf{x},t)$ into a structured latent representation $z_t$ that preserves multi-scale correlations. 
Temporal rollout is then performed in the latent space through numerical integration of a parameter-conditioned vector field learned from discrete observations, where physical parameters such as the viscosity coefficient $\nu$ and the numerical time step $\Delta t$ are embedded into the latent dynamics. This formulation enables generalization across different parameter configurations within the same class of PDEs. The latent vector field is learned via flow matching on latent state trajectories and supports continuous-time integration at arbitrary temporal query points. The single-step prediction procedure of AFNO is summarized as
\begin{equation}
u_t \xrightarrow{\mathrm{Enc}} z_t \xrightarrow{\text{latent flow}} z_{t+\Delta t} \xrightarrow{\mathrm{Dec}} \hat u_{t+\Delta t},
\end{equation}
where $\mathrm{Enc}$ and $\mathrm{Dec}$ denote the FNO-based encoder and decoder, respectively. The encoder, latent flow module, and decoder are trained jointly in an end-to-end manner to ensure consistent representation learning, temporal evolution, and reconstruction of the physical field.

\subsection{Latent Representation Module}
\label{sec:4.2}
Consider a solution field discretized on a regular grid and represented as $u_t\in\mathbb{R}^{C\times N_1\times\cdots\times N_d}$. Such data are inherently high-dimensional and exhibit strong multi-scale structure. Direct physical space modeling of temporal evolution is complicated by long-range spatial coupling, fine-scale local variations, and redundant degrees of freedom.
Such factors increase the sensitivity of rollout dynamics to approximation errors, particularly under long-horizon prediction. To address this issue, we introduce an FNO-based autoencoder (FNO-AE) as the latent representation module, which maps the physical field $u_t$ into a structured latent representation $z_t$. Rather than serving as a purely low-dimensional compression, the latent space reorganizes the original state into a compact and regular dynamical representation. This representation preserves dominant multi-scale correlations while filtering redundant variations that may amplify errors during physical space rollout.

FNO leverages spectral convolutions to directly learn mappings from input functions to output functions, performing global convolutions in Fourier space to capture multi-scale features. In the encoder, each Fourier layer in FNO is represented by an integral operator $\mathcal{K}$, consisting of stacked pointwise convolutions and Fourier operator modules, and expressed as:
\begin{equation}
(\mathcal{K}u)(x)=Wu(x)+\mathcal{F}^{-1}\left(R(k)\cdot\mathcal{F}[u](k)\right),
\end{equation}
where $\mathcal{F}$ denotes the $d$-dimensional discrete Fourier transform, $W$ is a pointwise linear operator corresponding to a $1\times 1$ convolution, and $R(k)$ is the learnable spectral kernel indexed by the multi-dimensional frequency $k$. In practice, high-frequency modes are truncated to focus learning on the dominant low-frequency components.

The encoder consists of stacked Fourier layers. For the input $h^{(0)}(x)=u_t(x)$, the output of the $(l+1)$-th layer is given by:
\begin{equation}
h^{(l+1)}(x)=\sigma\left(\mathcal{K}^{(l)}h^{(l)}(x)+B^{(l)}h^{(l)}(x)\right),
\end{equation}
where $\sigma(\cdot)$ denotes the nonlinear activation function, and $B^{(l)}$ denotes a channel-wise linear transformation acting pointwise in the spatial domain, which accounts for local interactions not captured by the truncated spectral convolution. After passing through $L$ Fourier layers, the encoder maps the input to the latent representation
\begin{equation}
z_t(x)=h^{(L)}(x)=\mathrm{Enc}(u_t),
\end{equation}
where $\mathrm{Enc}$ denotes the encoder mapping.

For the decoder, FNO establishes a consistent mapping between the spatial and spectral domains through the Fast Fourier Transform (FFT) \cite{nussbaumer1981fast} and the Inverse Fast Fourier Transform (IFFT), and the latent representation preserves the spectral structure of the underlying function. A symmetric architecture is therefore employed to reconstruct the physical field from the latent representation. Formally, the decoder is defined as
\begin{equation}
\hat u_t = \mathrm{Dec}(z_t),
\end{equation}
where $\mathrm{Dec}$ is implemented by a symmetric FNO-based architecture.

\subsection{Latent Flow Matching Module}
\label{sec:4.3}

AFNO replaces autoregressive rollout in the physical field space with autoregression-free rollout in the latent space.
In practice, the available observations are discrete physical-field samples, while latent states are accessible only through the encoder. As a consequence, temporal evolution in the latent space is not directly observable, and temporal derivatives are not directly available for supervision. This observation makes flow matching a suitable choice for learning latent dynamics from discrete PDE trajectories.
Here, flow matching is adapted to deterministic PDE evolution by using finite-difference increments between adjacent latent states as local velocity supervision, rather than modeling generic distributional transport paths.
This associates the learned vector field with the observed direction of latent evolution and supports numerical integration over long horizons.

For a given PDE $\partial_t u=\mathcal{N}(u;\nu)$, its temporal evolution forms a semigroup $\{\Phi_t\}_{t\ge 0}$ that satisfies $u(t+\Delta t)=\Phi_{\Delta t}(u(t))$. Denoting $z_t=\mathrm{Enc}(u_t)$, the corresponding latent evolution operator is defined as
\begin{equation}
\Psi_{\Delta t}=\mathrm{Enc}\circ\Phi_{\Delta t}\circ\mathrm{Dec},
\end{equation}
where $\mathrm{Dec}$ maps the latent representation back to the physical space.
Since $\Psi_{\Delta t}$ is not directly accessible from the observations, AFNO models the latent evolution as
\begin{equation}
\frac{dz}{dt}
=
v_{\theta}\bigl(z,b(\nu,\Delta t)\bigr),
\end{equation}
where $b(\nu,\Delta t)$ embeds the physical parameter $\nu$ and the time step $\Delta t$, and $v_{\theta}$ denotes the corresponding parameter-conditioned latent vector field.

Given the current latent state, temporal prediction is performed through numerical integration rather than recursive prediction in the physical field space. In this work, we employ the first-order explicit Euler update:
\begin{equation}
\begin{split}
\hat z_{t+\Delta t} &= \hat z_t+\Delta t\cdot v_\theta\big(\hat z_t,\,b(\nu,\Delta t)\big), \\
\hat u_{t+\Delta t} &= \mathrm{Dec}(\hat z_{t+\Delta t}).
\end{split}
\end{equation}
Multi-step prediction is therefore realized through iterative updates in the latent space, while the physical space prediction is recovered only after decoding.

Since only discrete latent states $\{z_t\}$ are available, we use finite-difference temporal derivatives constructed from adjacent latent states, namely
\begin{equation}
z_{t+\Delta t}-z_t\approx \Delta t\cdot v_\theta\big(z_t,b(\nu,\Delta t)\big),
\end{equation}
which leads to the flow matching loss:
\begin{equation}
\mathcal{L}_{\mathrm{flow}} =\mathbb{E}_{t}\left[ \left\| v_\theta\big(z_t,b(\nu,\Delta t)\big) -\frac{z_{t+\Delta t}-z_t}{\Delta t} \right\|_{L^2(\Omega)}^2 \right].
\end{equation}
Here, $\|\cdot\|_{L^2(\Omega)}$ denotes the $L^2$ norm over the spatial domain $\Omega$, and $\mathbb{E}_t[\cdot]$ represents the empirical expectation over sampled temporal indices along the training trajectories. In practice, the finite difference between $z_t=\mathrm{Enc}(u_t)$ and $z_{t+\Delta t}=\mathrm{Enc}(u_{t+\Delta t})$ serves as the supervision signal for learning the conditioned vector field, thereby aligning latent evolution with the observed trajectory increments.


\subsection{Physics Parameter Conditioning}
\label{sec:4.4}
Within a given PDE family, such as the viscous Burgers equations, the physical parameter $\nu$ and the numerical time step $\Delta t$ significantly influence the dynamical time scale and diffusion strength. To model different parameter configurations consistently within the same PDE class, a conditional embedding is defined as
\begin{equation}
b(\nu,\Delta t)=\mathrm{MLP}([\nu,\Delta t])\in\mathbb{R}^m,
\end{equation}
where MLP denotes a multilayer perceptron \cite{rumelhart1986learning}. This embedding is then spatially aligned to the dimensions $(N_1\times N_2\times\cdots\times N_d)$ of the latent space $z_t$. The aligned embedding $\tilde{b}$ is concatenated with the latent variable and fed into the vector field network:
\begin{equation}
v_\theta(z_t,b)=g_\theta\big([z_t,\tilde{b}]\big),
\end{equation}
where $g_\theta$ denotes the latent vector field network, which governs the parameter-conditioned temporal evolution in the latent space.

To capture long-range spatial dependencies and improve generalization across spatial locations, the vector field $v_\theta$ employs spectral convolutions with shared parameters in the frequency domain. Specifically, for the grid representation $z_t\in\mathbb{R}^{C_z\times N_1\times\cdots\times N_d}$, the network $g_\theta$ adopts a residual architecture composed of spectral and pointwise convolutions:
\begin{equation}
\begin{aligned}
h^{(0)} &= \mathrm{Conv}_{1\times 1}([z_t,\tilde{b}],k),\\
h^{(\ell+1)} &= \sigma\big(\mathrm{SpecConv}^{(\ell)}(h^{(\ell)})
+\mathrm{Conv}_{1\times 1}^{(\ell)}(h^{(\ell)})\big).
\end{aligned}
\end{equation}

Multiple residual layers are stacked to produce the vector field $v_\theta$.
The $\mathrm{SpecConv}$ operates in the $d$-dimensional Fourier domain with truncated low-frequency modes, enabling global spatial coupling and parameter sharing. This design supports adaptive latent dynamics across different parameter settings, including unseen values.

\begin{figure*}[h]
  \centering
  \includegraphics[width=0.9\linewidth]{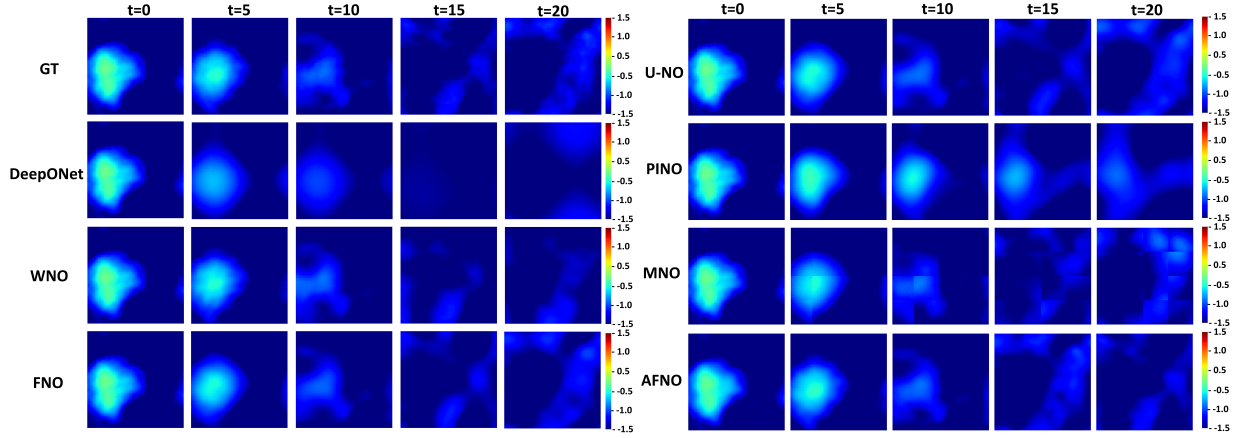}
   \caption{Qualitative comparison of predicted solutions over time for the 2D shallow water equation.}
  \label{fig:3}
\end{figure*}
\begin{figure*}[h]
  \centering
  \includegraphics[width=0.9\linewidth]{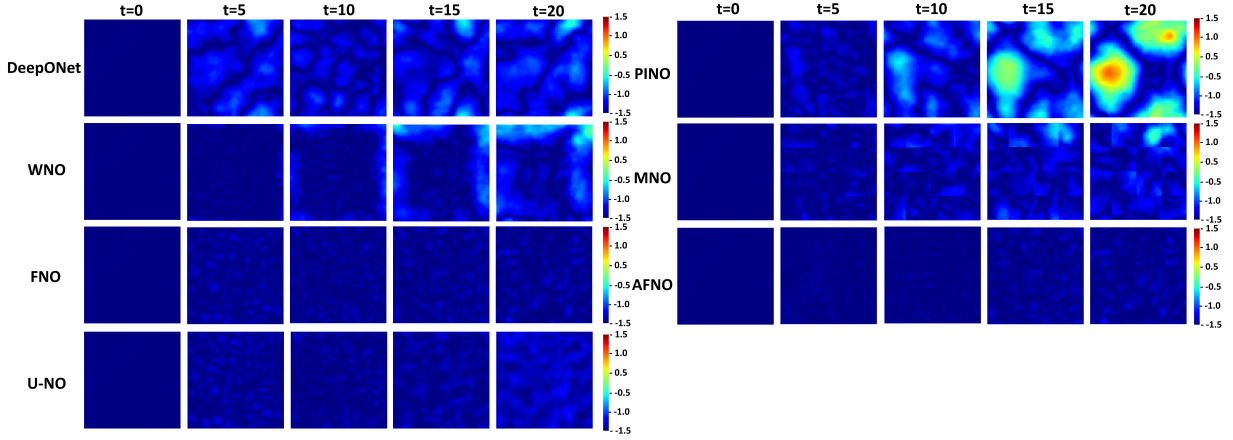}
   \caption{Qualitative comparison of error maps over time for the 2D shallow water equation.}
  \label{fig:11}
\end{figure*}
\subsection{Loss Functions}
\label{sec:4.5}
The training objective of AFNO combines three complementary terms: a reconstruction loss for preserving physical-field fidelity, a flow matching loss for learning latent dynamics, and a short-horizon rollout loss for constraining multi-step prediction.
The autoencoder reconstruction loss accounts for both field magnitude and gradient structure:
\begin{equation}
\mathcal{L}_{\mathrm{rec}} =
\|u_t-\hat u_t\|_{L^2(\Omega)}^2
+
\lambda_{H^1}\|\nabla u_t-\nabla \hat u_t\|_{L^2(\Omega)}^2,
\end{equation}
where $\lambda_{H^1}>0$ is a weighting coefficient that balances reconstruction fidelity and gradient consistency. 
Both terms are approximated on the discrete grid via weighted summation and central differences. 
The $H^1$ semi-norm term $\|\nabla u_t - \nabla \hat u_t\|_{L^2(\Omega)}^2$ encourages the latent representation to preserve first-order geometric structure.

The short-horizon rollout loss is defined by performing $K$ Euler updates in the latent space starting from $t_0$, where $s=0,1,\dots,K-1$ denotes the rollout step index:
\begin{equation}
\begin{aligned}
z^{(s+1)}
&=
z^{(s)}+\Delta t\cdot v_\theta\big(z^{(s)}, b(\nu,\Delta t)\big),\\
\hat u_{t_0+(s+1)\Delta t}
&=
\mathrm{Dec}\big(z^{(s+1)}\big).
\end{aligned}
\end{equation}
The corresponding loss is
\begin{equation}
\mathcal{L}_{\mathrm{roll}}
=
\mathbb{E}_{t_0,\nu,\Delta t}
\left[
\sum_{s=1}^{K}
\left\|
u_{t_0+s\Delta t}
-
\hat u_{t_0+s\Delta t}
\right\|_{L^2(\Omega)}^2
\right].
\end{equation}

This term constrains multi-step prediction errors and, together with $\mathcal{L}_{\mathrm{flow}}$ from Section~\ref{sec:4.3}, helps reduce error accumulation during rollout. 
The overall training objective is
\begin{equation}
\mathcal{L}
=
\lambda_{\mathrm{rec}}\mathcal{L}_{\mathrm{rec}}
+
\lambda_{\mathrm{flow}}\mathcal{L}_{\mathrm{flow}}
+
\lambda_{\mathrm{roll}}\mathcal{L}_{\mathrm{roll}}.
\end{equation}

\section{Experiments}
\label{sec:5}
In this section, we present experimental results on benchmark PDEs, including error metrics across different equations, long-horizon rollout evaluations, and related analyses.

\subsection{Setups}
\label{sec:5.1}

\noindent\textbf{Benchmarks.}
For a comprehensive and reliable evaluation, we conduct experiments on the public benchmark PDEBench \cite{takamoto2022pdebench}. All samples are generated following the PDEBench data generation protocol under numerically stable settings. On this basis, the benchmark suite is further extended to include additional time-dependent PDE types. During data generation, numerical solvers are selected according to the physical characteristics of each PDE. For instance, the 2D Navier-Stokes equations are solved using a pseudo-spectral method \cite{hou2007computing} with Crank-Nicolson time integration \cite{crank1947practical}, and de-aliasing techniques are employed to suppress high-frequency numerical artifacts. For 1D benchmarks, we consider the 1D-Burgers equation, the 1D Allen-Cahn (1D-AC) equation, and the 1D Kuramoto-Sivashinsky (1D-KS) equation. For 2D benchmarks, we evaluate performance on the 2D Navier-Stokes (2D-NS) equations, the 2D Shallow Water Equations (2D-SWE), and the 2D complex Ginzburg-Landau (2D-CGL) equation.

\noindent\textbf{Baseline Methods.}
AFNO is compared with six representative baseline models, including the FNO \cite{li2020fourier}, DeepONet \cite{lu2019deeponet}, U-shaped Neural Operator (U-NO) \cite{rahman2022u}, Physics-Informed Neural Operator (PINO) \cite{li2024physics}, Wavelet Neural Operator (WNO) \cite{tripura2023wavelet}, and Mamba Neural Operator (MNO) \cite{cheng2025mamba}. In addition to the above neural operators, we further compare with several recently proposed models specifically developed to address long-horizon forecasting and error accumulation in autoregressive operator learning, including RNO \cite{ye2025recurrent}, LITEFNO \cite{ahn2025lightweight}, and FreqMoE \cite{chen2025freqmoe}. Together, these baselines provide a comprehensive comparison across classical neural operators, recent operator architectures, and methods specifically designed for stable long-horizon extrapolation.
\begin{figure*}[t]
  \centering
  \includegraphics[width=0.9\linewidth]{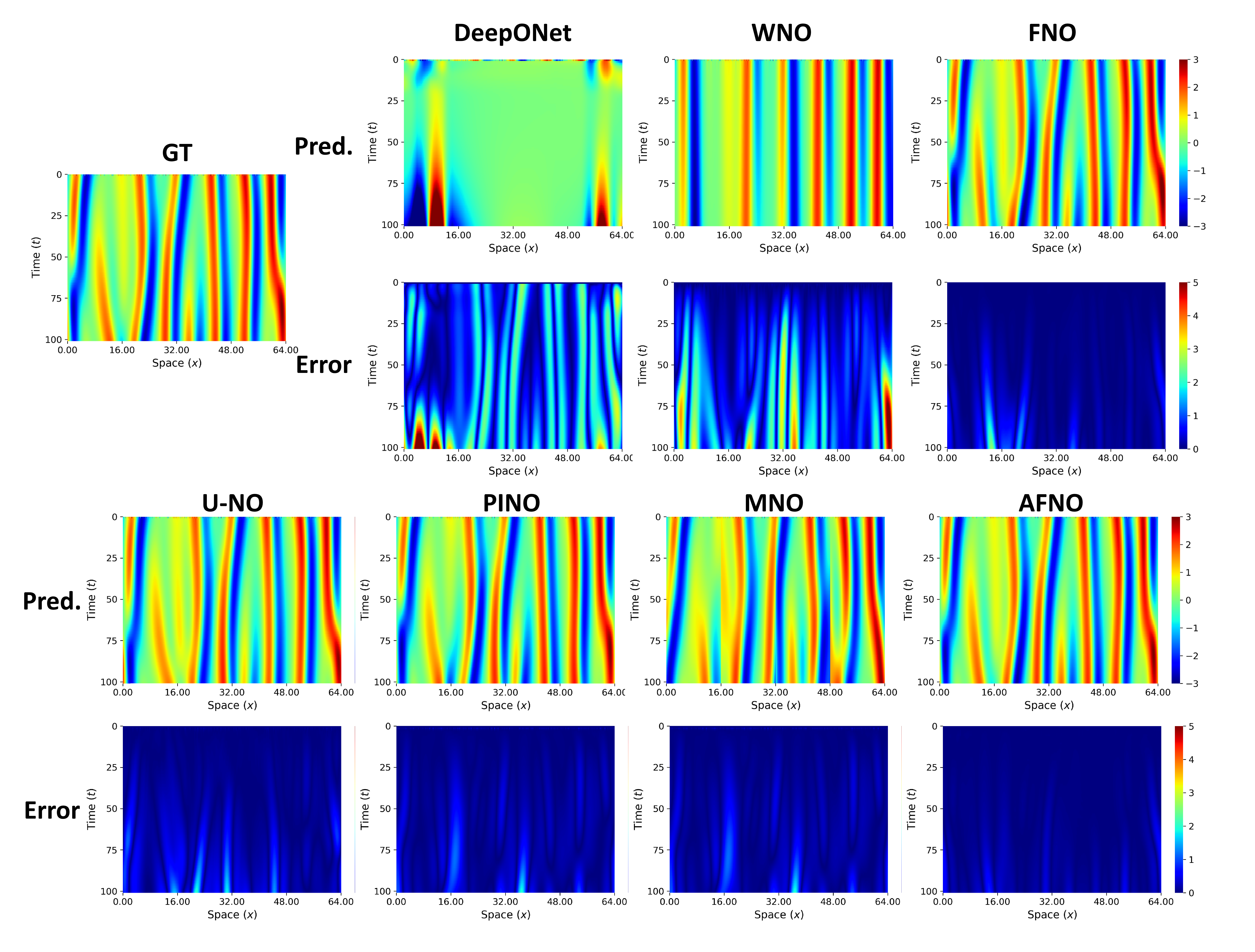}
   \caption{Comparison of time-evolving predicted solutions and corresponding error maps for the 1D Kuramoto-Sivashinsky equation.}
  \label{fig:4}
\end{figure*}

\noindent\textbf{Implementation Details.}
All experiments are conducted on a single NVIDIA RTX 4090 GPU with 48 GB memory using the PyTorch~2.5.1 framework. The training set contains 1,000 samples, and the test set consists of 200 independently generated trajectories. MSE is adopted as the training objective for baseline models. Models are optimized using the Adam \cite{kingma2014adam} optimizer with an initial learning rate of $0.001$, together with a cosine annealing scheduler for learning rate decay. Each model is trained for 500 epochs to ensure convergence. The batch size is 32 for 1D PDEs and 8 for 2D PDEs. AFNO uses $C_z=16$, hidden width 64, and $k=4$ Fourier modes and all models employ a pushforward training strategy with a discrete time horizon of 5 \cite{brandstetter2022message}. The loss weights are set to $\lambda_{\mathrm{rec}}=1$, $\lambda_{\mathrm{flow}}=\lambda_{\mathrm{roll}}=0$ for the first 30\% epochs, and to (0, 0.5, 0.1) thereafter. To ensure resolution invariance, training inputs are downsampled to a fixed resolution (256 for 1D PDEs and $64 \times 64$ for 2D PDEs). During inference, inputs are first downsampled to the training resolution, processed by the model, and then upsampled back to the original resolution. 

\noindent\textbf{Evaluation Metrics.}
Model performance is evaluated using a comprehensive set of quantitative metrics. In addition to standard relative error measures, absolute error metrics are included to quantify deviations. All metrics are averaged over the test set. Specifically, we report the relative $L_2$ error (Rel.$L_2$), relative $H_1$ error (Rel.$H_1$), mean squared error (MSE), root mean squared error (RMSE), mean absolute error (MAE), maximum absolute error (MaxError), normalized RMSE (nRMSE), final-time RMSE (fRMSE), and cumulative RMSE (cRMSE). To further evaluate performance and stability under long-horizon temporal rollouts, we additionally examine the evolution of MSE, Rel.$L_2$, and Rel.$H_1$ over $T$ rollout steps, which characterizes the accumulation of prediction errors over time. Unless otherwise specified, these rollout errors are computed at each rollout step and averaged over the test set. 

\begin{table*}[t]
\centering
\caption{Performance comparison of different models across multiple PDEs, evaluated using the trained models at the end of training. \textbf{Bold} values with \colorbox{yellow!30}{yellow background} indicate the best performance for each metric.}
\label{tab:main_results}
\resizebox{\textwidth}{!}{
\begin{tabular}{llccccccccc}
\toprule
Equation & Model & Rel.$L_2$ & Rel.$H_1$ & MSE & RMSE & MAE & MaxError & nRMSE & fRMSE & cRMSE \\
\midrule
\multirow{7}{*}{1D-Burgers} 
& DeepONet & 4.23E-02 & 2.09E-01 & 1.04E-13 & 3.19E-07 & 2.56E-07 & 7.90E-07 & 1.35E-02 & 2.74E+04 & 8.99E+04 \\
& WNO & 1.25E-02 & 1.61E-01 & 8.91E-15 & 9.38E-08 & 7.40E-08 & 2.82E-07 & 4.00E-03 & 4.53E-06 & 2.19E-04 \\
& FNO & 1.13E-02 & 1.66E-02 & 8.71E-15 & 9.11E-08 & 7.54E-08 & 1.76E-07 & 3.66E-03 & 2.80E-06 & 1.41E-04 \\
& U-NO & 1.69E-03 & 1.22E-02 & 2.71E-16 & 1.57E-08 & 1.21E-08 & 3.50E-08 & 5.47E-04 & \cellcolor{yellow!30}\textbf{3.63E-07} & 1.91E-05 \\
& PINO & 9.53E-04 & 9.56E-03 & 1.20E-16 & 9.77E-09 & 7.23E-09 & 2.24E-08 & 3.10E-04 & 7.15E-07 & 3.64E-05 \\
& MNO & 3.71E-02 & 1.67E+00 & 7.69E-14 & 2.77E-07 & 2.07E-07 & 1.12E-06 & 1.17E-02 & 1.10E-06 & 7.33E-05 \\
& AFNO & \cellcolor{yellow!30}\textbf{2.23E-04} & \cellcolor{yellow!30}\textbf{1.13E-03} & \cellcolor{yellow!30}\textbf{2.82E-18} & \cellcolor{yellow!30}\textbf{1.68E-09} & \cellcolor{yellow!30}\textbf{1.33E-09} & \cellcolor{yellow!30}\textbf{4.10E-09} & \cellcolor{yellow!30}\textbf{7.12E-05} & 1.27E-07 & \cellcolor{yellow!30}\textbf{1.07E-05} \\
\midrule
\multirow{7}{*}{1D-AC} 
& DeepONet & 1.85E-01 & 5.06E-01 & 1.37E-08 & 1.16E-04 & 5.53E-05 & 1.29E-04 & 5.95E-02 & 1.40E-03 & 2.64E-02 \\
& WNO & 1.86E-02 & 3.27E-02 & 1.85E-09 & 4.05E-05 & 9.33E-06 & 3.11E-05 & 6.15E-03 & 1.85E+00 & 7.15E+00 \\
& FNO & 6.85E-02 & 1.26E-01 & 1.60E-08 & 1.24E-04 & 4.07E-05 & 1.17E-04 & 2.27E-02 & 7.70E-03 & 1.46E-01 \\
& U-NO & 2.40E-01 & 2.86E-01 & 7.15E-08 & 2.63E-04 & 1.12E-04 & 2.42E-04 & 7.94E-02 & 1.39E-03 & 2.58E-02 \\
& PINO & 6.66E-02 & 1.44E-01 & 3.88E-09 & 6.14E-05 & 2.66E-05 & 6.94E-05 & 2.18E-02 & 5.08E-02 & 3.53E-01 \\
& MNO & 1.16E-01 & 8.33E+00 & 7.19E-09 & 8.34E-05 & 2.95E-05 & 2.42E-04 & 3.83E-02 & 3.82E-03 & 6.67E-02 \\
& AFNO & \cellcolor{yellow!30}\textbf{5.98E-03} & \cellcolor{yellow!30}\textbf{1.59E-02} & \cellcolor{yellow!30}\textbf{2.33E-12} & \cellcolor{yellow!30}\textbf{1.51E-06} & \cellcolor{yellow!30}\textbf{7.15E-07} & \cellcolor{yellow!30}\textbf{1.51E-06} & \cellcolor{yellow!30}\textbf{1.90E-03} & \cellcolor{yellow!30}\textbf{4.88E-04} & \cellcolor{yellow!30}\textbf{9.85E-03} \\
\midrule
\multirow{7}{*}{1D-KS} 
& DeepONet & 8.37E-01 & 9.09E-01 & 1.21E+00 & 1.10E+00 & 8.70E-01 & 2.60E+00 & 2.16E-01 & 2.22E+01 & 5.39E+02 \\
& WNO & 6.25E-02 & 4.34E-01 & 6.67E-03 & 8.16E-02 & 6.47E-02 & 2.97E-01 & 1.61E-02 & 1.53E+00 & 1.04E+02 \\
& FNO & 9.93E-03 & 5.51E-02 & \cellcolor{yellow!30}\textbf{1.64E-04} & \cellcolor{yellow!30}\textbf{1.28E-02} & \cellcolor{yellow!30}\textbf{9.79E-03} & 6.04E-02 & \cellcolor{yellow!30}\textbf{2.51E-03} & 4.41E-01 & 1.76E+01 \\
& U-NO & 5.41E-02 & 5.15E-01 & 4.96E-03 & 7.04E-02 & 5.43E-02 & 2.83E-01 & 1.39E-02 & 7.50E-01 & 3.06E+01 \\
& PINO & 1.09E-01 & 8.81E-01 & 2.01E-02 & 1.42E-01 & 1.12E-01 & 5.01E-01 & 2.79E-02 & 7.98E-01 & 3.57E+01 \\
& MNO & 4.05E-02 & 3.50E-01 & 2.80E-03 & 5.29E-02 & 4.20E-02 & 1.90E-01 & 1.04E-02 & 1.50E+00 & 6.82E+01 \\
& AFNO & \cellcolor{yellow!30}\textbf{9.78E-03} & \cellcolor{yellow!30}\textbf{1.41E-02} & 1.69E-04 & 1.30E-02 & 1.08E-02 & \cellcolor{yellow!30}\textbf{3.28E-02} & 2.55E-03 & \cellcolor{yellow!30}\textbf{2.32E-01} & \cellcolor{yellow!30}\textbf{9.68E+00} \\
\midrule
\multirow{7}{*}{2D-NS} 
& DeepONet & 8.62E-02 & 1.45E-01 & 5.60E-03 & 7.49E-02 & 4.74E-02 & 1.95E-01 & 3.05E-02 & 1.36E-01 & 1.68E+00 \\
& WNO & 4.83E-03 & 2.05E-02 & 3.56E-06 & 1.89E-03 & 1.50E-03 & 8.92E-03 & 1.71E-03 & 7.32E-02 & 6.44E-01 \\
& FNO & 2.93E-03 & 6.26E-03 & \cellcolor{yellow!30}\textbf{2.05E-07} & \cellcolor{yellow!30}\textbf{4.52E-04} & \cellcolor{yellow!30}\textbf{3.28E-04} & \cellcolor{yellow!30}\textbf{2.13E-03} & \cellcolor{yellow!30}\textbf{3.54E-04} & 1.97E-02 & 1.80E-01 \\
& U-NO & 1.09E-03 & 6.87E-03 & 2.44E-07 & 4.93E-04 & 3.61E-04 & 2.41E-03 & 3.85E-04 & 4.07E-03 & 3.81E-02 \\
& PINO & 3.44E-03 & 4.93E-03 & 2.22E-06 & 1.49E-03 & 1.14E-03 & 3.00E-03 & 1.22E-03 & 9.96E-03 & 1.64E-01 \\
& MNO & 5.56E-02 & 3.99E-01 & 2.86E-05 & 5.34E-03 & 4.20E-03 & 2.62E-02 & 1.96E-02 & 2.31E-03 & 5.50E-02 \\
& AFNO & \cellcolor{yellow!30}\textbf{1.00E-03} & \cellcolor{yellow!30}\textbf{1.82E-03} & 1.77E-06 & 1.33E-03 & 1.15E-03 & 2.11E-03 & 1.04E-03 & \cellcolor{yellow!30}\textbf{1.60E-03} & \cellcolor{yellow!30}\textbf{2.97E-02} \\
\midrule
\multirow{7}{*}{2D-SWE} 
& DeepONet & 5.60E-02 & 1.37E-01 & 1.13E-03 & 3.35E-02 & 2.53E-02 & 1.15E-01 & 2.55E-02 & 2.74E-02 & 5.43E-01 \\
& WNO & 4.81E-03 & 3.63E-02 & 8.35E-06 & 2.88E-03 & 2.17E-03 & 1.41E-02 & 2.19E-03 & 3.40E-02 & 3.27E-01 \\
& FNO & 7.63E-03 & 5.19E-02 & 2.12E-05 & 4.59E-03 & 3.47E-03 & 2.06E-02 & 3.47E-03 & 6.41E-03 & 1.07E-01 \\
& U-NO & 6.57E-03 & 4.84E-02 & 1.57E-05 & 3.94E-03 & 2.95E-03 & 1.94E-02 & 2.99E-03 & 1.16E-02 & 1.22E-01 \\
& PINO & 1.67E-02 & 5.93E-02 & 9.72E-05 & 9.83E-03 & 7.52E-03 & 4.10E-02 & 7.62E-03 & 6.77E-02 & 8.40E-01 \\
& MNO & 7.42E-03 & 6.28E-02 & 1.98E-05 & 4.43E-03 & 3.31E-03 & 2.59E-02 & 3.38E-03 & 2.99E-02 & 3.33E-01 \\
& AFNO & \cellcolor{yellow!30}\textbf{2.30E-03} & \cellcolor{yellow!30}\textbf{2.28E-02} & \cellcolor{yellow!30}\textbf{1.91E-06} & \cellcolor{yellow!30}\textbf{1.38E-03} & \cellcolor{yellow!30}\textbf{1.00E-03} & \cellcolor{yellow!30}\textbf{9.63E-03} & \cellcolor{yellow!30}\textbf{1.05E-03} & \cellcolor{yellow!30}\textbf{3.67E-03} & \cellcolor{yellow!30}\textbf{5.64E-02} \\
\midrule
\multirow{7}{*}{2D-CGL} 
& DeepONet & 7.62E-01 & 8.01E-01 & 1.37E-01 & 3.70E-01 & 2.95E-01 & 1.52E+00 & 1.41E-01 & 5.94E-01 & 9.49E+00 \\
& WNO & 8.00E-03 & 9.87E-03 & 1.56E-05 & 3.95E-03 & 3.05E-03 & 4.11E-02 & 1.47E-03 & 7.59E-02 & 6.29E-01 \\
& FNO & 2.15E-02 & 2.57E-02 & 1.14E-04 & 1.07E-02 & 7.99E-03 & 6.85E-02 & 3.95E-03 & 2.16E-01 & 2.07E+00 \\
& U-NO & 1.07E-02 & 1.34E-02 & 2.87E-05 & 5.36E-03 & 4.05E-03 & 3.31E-02 & 1.98E-03 & 9.01E-02 & 8.77E-01 \\
& PINO & 1.19E-02 & 1.36E-02 & 3.43E-05 & 5.85E-03 & 4.46E-03 & 3.44E-02 & 2.19E-03 & 8.70E-02 & 8.46E-01 \\
& MNO & 1.78E-02 & 3.11E-02 & 7.72E-05 & 8.79E-03 & 6.53E-03 & 7.55E-02 & 3.28E-03 & 1.26E-01 & 1.29E+00 \\
& AFNO & \cellcolor{yellow!30}\textbf{5.66E-03} & \cellcolor{yellow!30}\textbf{7.47E-03} & \cellcolor{yellow!30}\textbf{7.91E-06} & \cellcolor{yellow!30}\textbf{2.81E-03} & \cellcolor{yellow!30}\textbf{1.77E-03} & \cellcolor{yellow!30}\textbf{2.14E-02} & \cellcolor{yellow!30}\textbf{1.04E-03} & \cellcolor{yellow!30}\textbf{4.17E-02} & \cellcolor{yellow!30}\textbf{4.37E-01} \\
\bottomrule
\end{tabular}}
\end{table*}

\subsection{Main Results}
\label{sec:5.2}
As shown in Table~\ref{tab:main_results}, AFNO is systematically compared with several representative neural operator methods across multiple 1D and 2D PDE benchmarks. AFNO exhibits lower prediction errors on most benchmarks, particularly in terms of the Rel.$L_2$ and Rel.$H_1$ metrics. Although Rel.$L_2$ measures the global reconstruction error relative to the reference energy and Rel.$H_1$ further reflects the consistency of spatial derivatives, these results indicate that AFNO accurately recovers both the overall field distribution and its structure of local variation. For the 1D benchmarks, AFNO reports the lowest relative errors on the 1D-Burgers and 1D-AC equations among the compared methods. On the Burgers equation, the Rel.$L_2$ is $2.23\times10^{-4}$ and the Rel.$H_1$ is $1.13\times10^{-3}$. In addition, AFNO also performs favorably on metrics such as MSE, RMSE, and cRMSE, further demonstrating accurate pointwise prediction and stable temporal evolution. In the AC equation, the relative errors remain in the $10^{-3}$ order of magnitude, while consistently low values are observed for the metrics based on MSE, RMSE, and cRMSE. For the 2D benchmarks, comparable trends are observed. On the 2D-NS equations and 2D-SWE, AFNO achieves Rel.$L_2$ and Rel.$H_1$ that rank among the lowest values reported in the table. On the 1D-KS equation, the advantage is smaller on certain pointwise error metrics, which is consistent with the chaotic and multi-scale nature of KS dynamics. AFNO nevertheless achieves lower Rel.$H_1$ error and better long-horizon performance.

Figs.~\ref{fig:3}, \ref{fig:11}, and~\ref{fig:4} provide qualitative examples for selected 2D and 1D benchmarks, including predicted solution fields and corresponding error maps over time. These examples correspond to a subset of the visualization results reported in this paper, while additional qualitative results for the other benchmarks and rollout settings are provided in the supplementary material. In the shown cases, AFNO is closer to the ground-truth evolution and presents more localized error patterns than the compared baselines. 


\begin{figure*}[t]
  \centering
  \includegraphics[width=\linewidth]{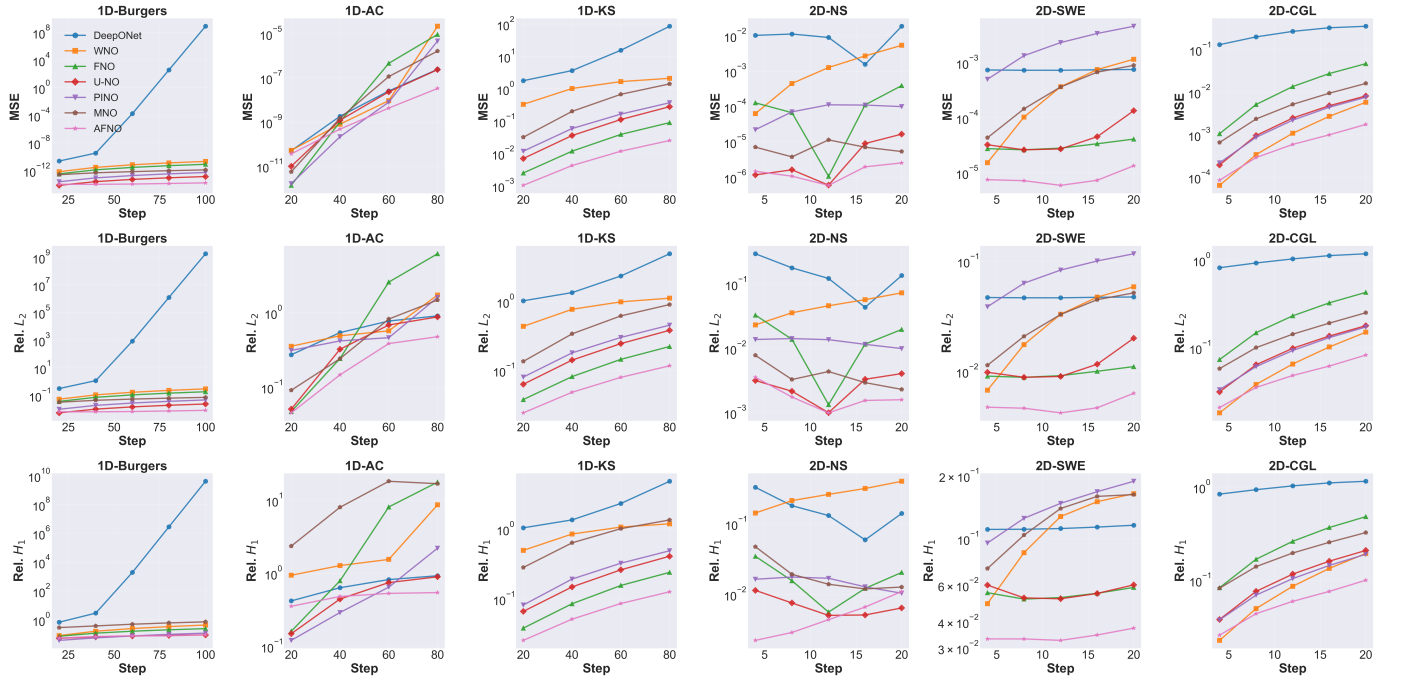}
   \caption{Rollout error evolution under extrapolation across the PDEs for different models.}
  \label{fig:2}
\end{figure*}

\begin{figure*}[t]
  \centering
  \includegraphics[width=0.85\linewidth]{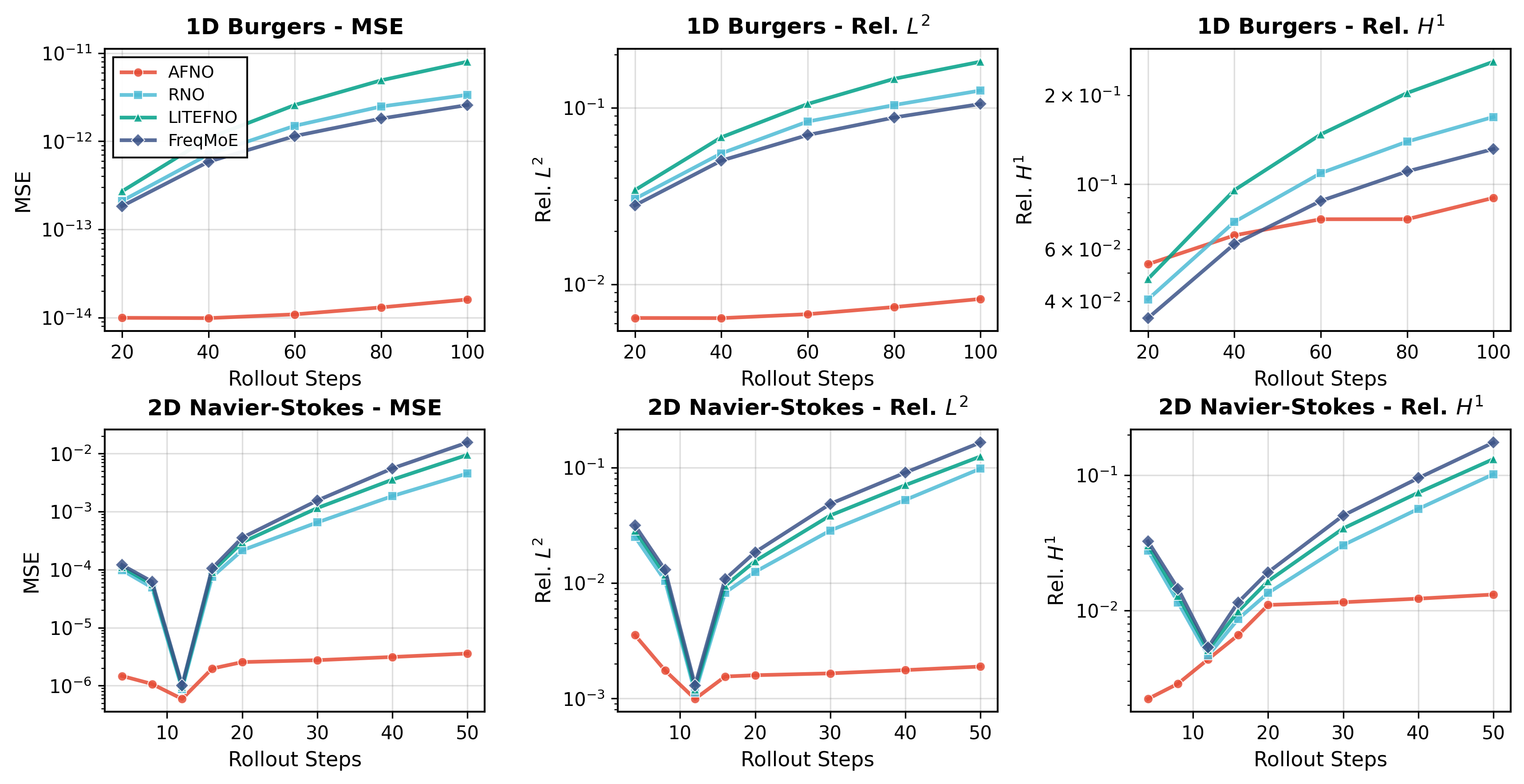}
   \caption{Comparison with long-horizon forecasting methods in the 1D-Burgers equation and 2D-NS equation.}
  \label{fig:long}
\end{figure*}

\subsection{Long-Horizon Rollout Evaluation}
\label{sec:5.3}
\textbf{Comparison with classical baselines.}
As shown in Fig.~\ref{fig:2}, AFNO exhibits slower error accumulation over long-horizon rollout across most benchmarks. In terms of the Rel.$L_2$ and Rel.$H_1$ metrics, AFNO shows slower and approximately linear error growth as the rollout horizon increases. For the 1D benchmarks, on the 1D-Burgers equation, the MSE of AFNO remains below the $10^{-2}$ level after 100 rollout steps, whereas other methods, such as MNO and FNO, exhibit faster error accumulation. In the 1D-AC equation, the relative error curves are similarly smooth, with Rel.$L_2$ and Rel.$H_1$ maintaining consistently low values over multiple rollout steps. Similar trends are observed on the 2D benchmarks, where AFNO continues to show slower error growth as the rollout horizon increases. In some cases, the final MSE and relative errors are reduced by more than 50\% compared to competing methods. On the 1D-KS equation, the advantage is smaller on certain pointwise error metrics, which is consistent with the chaotic and multi-scale nature of 1D-KS. AFNO nevertheless attains lower Rel.$H_1$ error and more favorable long-horizon metrics.

\textbf{Comparison with long-horizon forecasting methods.} We further compare AFNO with representative methods designed for long-horizon extrapolation, as shown in Fig.~\ref{fig:long}. On the 1D-Burgers equation, AFNO maintains the lowest rollout errors over 100 prediction steps. At the final step, its MSE remains on the order of $10^{-14}$, about two orders of magnitude smaller than the competing methods. The Rel.$L_2$ error remains below $10^{-2}$ for AFNO, whereas the baselines reach the $10^{-1}$ level, and Rel.$H^1$ increases more gradually throughout the rollout. A similar pattern is observed on the 2D-NS equation. At 50 rollout steps, AFNO keeps the MSE at the $10^{-6}$ level, while the competing methods have increased to around $10^{-3}$-$10^{-2}$, and its relative errors remain substantially lower over the entire rollout horizon. These results indicate that AFNO yields slower error accumulation and more stable long-horizon prediction than methods specifically developed for long-range forecasting, which is consistent with the error propagation behavior illustrated in Fig.~\ref{fig:case}.

\begin{figure}[t]
  \centering
  \includegraphics[width=\linewidth]{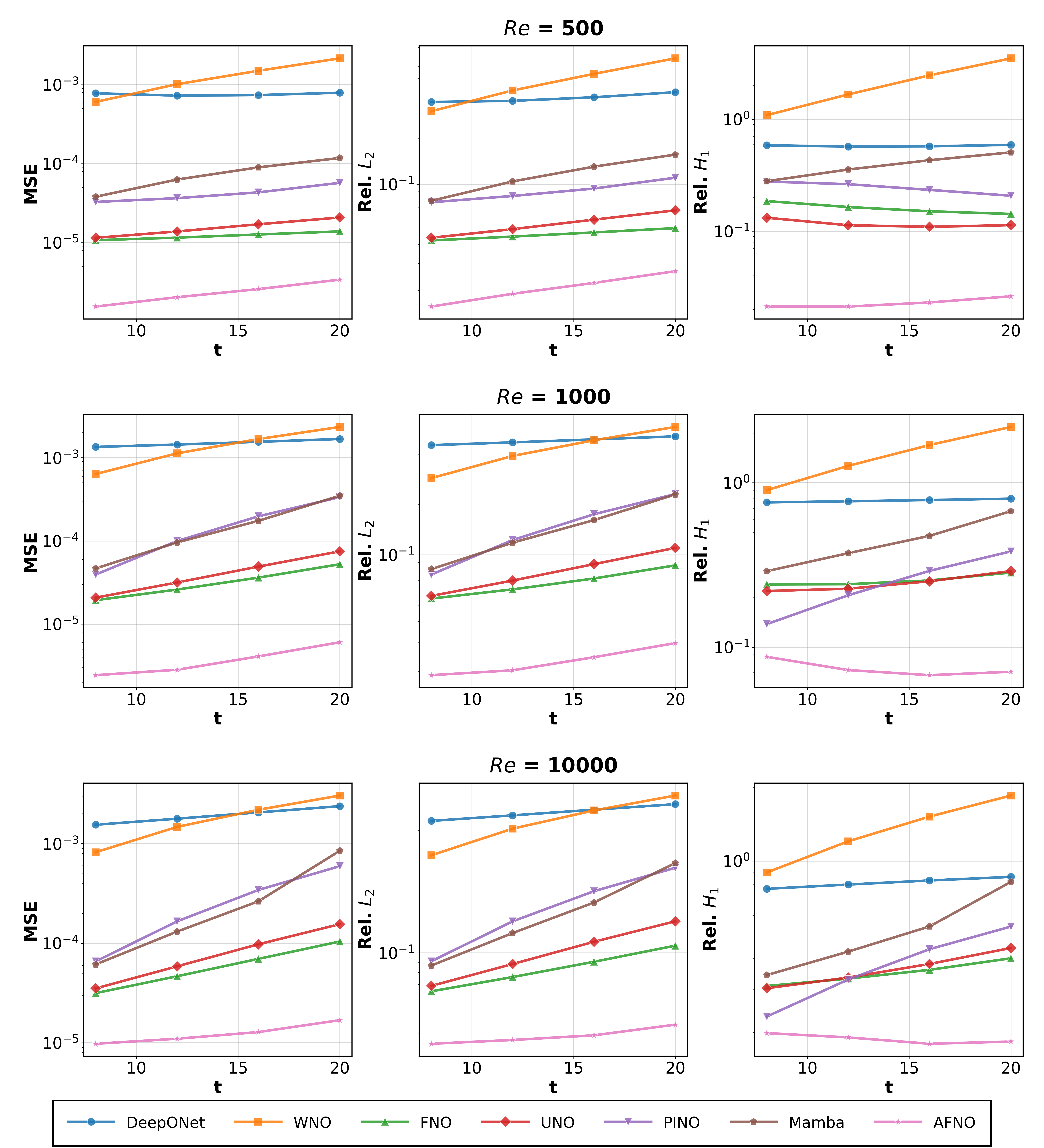}
   \caption{Temporal evolution of rollout errors for three metrics under cross-parameter evaluation.}
  \label{fig:5}
\end{figure}

\subsection{Cross-Parameter Evaluation}
\label{sec:5.4}
This section evaluates the physics parameter conditioning capability of AFNO across different parameter configurations within the same class of PDEs. Specifically, all models are trained on the 2D-NS equations under the parameter setting of $Re = 100$, and subsequently evaluated on previously unseen Reynolds. Table~\ref{tab:cross} reports the quantitative results at 20 rollout steps under different Reynolds ($Re$) number conditions. AFNO consistently achieves the best or substantially superior performance in all test settings in terms of MSE, Rel.$L_2$ and Rel.$H_1$. In contrast, other methods exhibit varying degrees of performance degradation as the parameters deviate from the training distribution, with the degradation being particularly pronounced in Rel.$H_1$, since the metric is sensitive to gradient structures and small-scale variation of the solution. 

\begin{table}[h]
\centering
\setlength{\tabcolsep}{12pt}
\caption{Cross-parameter generalization performance on 2D-NS at different Reynolds numbers.}
\label{tab:cross}
\begin{tabular}{lccc}
\toprule
Model & MSE & Rel.$L_2$ & Rel.$H_1$ \\
\midrule
\multicolumn{4}{c}{\textbf{$Re$ = 500}} \\
\midrule
DeepONet & 7.90E-04 & 4.04E-01 & 5.91E-01 \\
WNO & 2.16E-03 & 6.76E-01 & 3.50E+00 \\
FNO & 1.39E-05 & 5.14E-02 & 1.43E-01 \\
U-NO & 2.08E-05 & 6.73E-02 & 1.13E-01 \\
PINO & 5.71E-05 & 1.10E-01 & 2.08E-01 \\
MNO & 1.18E-04 & 1.57E-01 & 5.06E-01 \\
AFNO & \cellcolor{green!20}\textbf{3.39E-06} & \cellcolor{green!20}\textbf{2.67E-02} & \cellcolor{green!20}\textbf{2.62E-02} \\
\midrule
\multicolumn{4}{c}{\textbf{$Re$ = 1000}} \\
\midrule
DeepONet & 1.68E-03 & 5.12E-01 & 7.99E-01 \\
WNO & 2.34E-03 & 5.83E-01 & 2.18E+00 \\
FNO & 5.24E-05 & 8.65E-02 & 2.84E-01 \\
U-NO & 7.49E-05 & 1.10E-01 & 2.90E-01 \\
PINO & 3.36E-04 & 2.31E-01 & 3.82E-01 \\
MNO & 3.49E-04 & 2.30E-01 & 6.71E-01 \\
AFNO & \cellcolor{green!20}\textbf{6.04E-06} & \cellcolor{green!20}\textbf{2.96E-02} & \cellcolor{green!20}\textbf{7.09E-02} \\
\midrule
\multicolumn{4}{c}{\textbf{$Re$ = 10000}} \\
\midrule
DeepONet & 2.38E-03 & 5.40E-01 & 8.61E-01 \\
WNO & 3.04E-03 & 5.95E-01 & 1.85E+00 \\
FNO & 1.04E-04 & 1.09E-01 & 4.01E-01 \\
U-NO & 1.56E-04 & 1.43E-01 & 4.42E-01 \\
PINO & 5.94E-04 & 2.63E-01 & 5.40E-01 \\
MNO & 8.48E-04 & 2.77E-01 & 8.21E-01 \\
AFNO & \cellcolor{green!20}\textbf{1.69E-05} & \cellcolor{green!20}\textbf{4.43E-02} & \cellcolor{green!20}\textbf{1.83E-01} \\
\bottomrule
\end{tabular}
\end{table}

In addition, Fig.~\ref{fig:5} presents a comparison of error variation over multiple rollout steps under different physical parameter configurations. As the extrapolation horizon increases, AFNO maintains consistently lower error levels and exhibits a more stable error evolution across all test parameters. These results indicate that explicit parameter conditioning enables the model to preserve parameter-consistent dynamical behavior during iterative prediction, thereby substantially improving the stability and robustness of long-horizon extrapolation across varying physical parameters.
\begin{table*}[t]
\centering
\caption{Sensitivity analysis of loss weighting strategies on 1D-Burgers and 2D-NS equations}
\label{tab:sensitivity}
\begin{tabular}{ccccccccc}
\toprule
\multicolumn{3}{c}{Stage 2 Loss Weights} & \multicolumn{3}{c}{1D-Burgers} & \multicolumn{3}{c}{2D-NS} \\
\cmidrule(lr){1-3} \cmidrule(lr){4-6} \cmidrule(lr){7-9}
$\mathcal{L}_{\text{rec}}$ & $\mathcal{L}_{\text{flow}}$ & $\mathcal{L}_{\text{roll}}$ & MSE & Rel.$L^2$ & Rel.$H^1$ & MSE & Rel.$L^2$ & Rel.$H^1$ \\
\midrule
0.4 & 0.5 & 0.1 & 1.27E-14 & 1.47E-02 & 8.29E-02 & 2.51E-06 & 1.57E-03 & 1.07E-02 \\
1/3 & 1/3 & 1/3 & 2.01E-14 & 1.59E-02 & 6.16E-02 & 1.03E-06 & 1.01E-03 & 1.49E-03 \\
0.1 & 0.8 & 0.1 & 2.36E-14 & 1.86E-02 & 9.48E-02 & 1.33E-05 & 3.63E-03 & 3.37E-03 \\
0.1 & 0.1 & 0.8 & 6.66E-15 & 1.01E-02 & 6.43E-02 & 5.10E-06 & 2.25E-03 & 1.14E-02 \\
\bottomrule
\end{tabular}
\end{table*}

\begin{table}[h]
\centering
\caption{Architecture variants of AFNO in the ablation study}
\label{tab:architecture}
\begin{tabular}{p{1.5cm}p{2.5cm}p{3.4cm}}
\toprule
Variant & Modification & Description \\
\midrule
AFNO & - & Baseline model \\
AFNO-Conv & Latent Flow Matching Module & Replaced with 2-layer $3\times3$ local convolution \\
AFNO-MLP & Latent Flow Matching Module & Simplified to 2-layer $1\times1$ pointwise convolution \\
AFNO-Res & Encoder/Decoder & Replaced with ResNet (4 blocks, 8 layers $3\times3$) \\
AFNO-AR & Time evolution mechanism & Replaced with discrete autoregressive latent rollout \\
\bottomrule
\end{tabular}
\end{table}

\subsection{Ablation Study}
\label{sec:5.5}

To assess the contribution of individual components in AFNO, Table~\ref{tab:architecture} summarizes the architectural variants considered in the ablation study, with modifications to the latent flow matching module and the encoder-decoder architecture. Specifically, \textbf{AFNO-Conv} and \textbf{AFNO-MLP} replace the latent flow matching module with a local convolutional operator and a pointwise mapping, respectively. The \textbf{AFNO-Conv} variant examines the role of parameter conditioning as well as global interactions in latent evolution. By contrast, \textbf{AFNO-MLP} treats the latent representation as a spatially unstructured vector, allowing us to assess the importance of preserving spatial structure in the latent space. In addition, \textbf{AFNO-Res} replaces the operator-based encoder and decoder with a ResNet \cite{he2016deep} to evaluate the effect of the FNO-based design.

\begin{table}[h]
\centering
\caption{Ablation study comparison of AFNO variants}
\label{tab:abtion}
\begin{tabular}{llccc}
\toprule
Equation & Model & MSE & Rel.$L_2$ & Rel.$H_1$ \\
\midrule
\multirow{4}{*}{\shortstack{1D-Burgers}} 
& AFNO-Conv & 2.79E-13 & 6.98E-02 & 2.72E-01 \\
& AFNO-Res & 8.75E-14 & 3.60E-02 & 1.03E+00 \\
& AFNO-MLP & 3.89E-13 & 8.52E-02 & 4.24E-01 \\
& AFNO-AR & 9.75E-14 & 5.12E-02 & 8.25E-01 \\
& \cellcolor{cyan!15}\textbf{AFNO} & \cellcolor{cyan!15}\textbf{1.33E-14} & \cellcolor{cyan!15}\textbf{1.60E-02} & \cellcolor{cyan!15}\textbf{1.71E-01} \\
\midrule
\multirow{4}{*}{\shortstack{2D-NS}} 
& AFNO-Conv & 1.24E-04 & 1.11E-02 & 8.26E-03 \\
& AFNO-Res & 1.35E-05 & 3.65E-03 & 4.71E-03 \\
& AFNO-MLP & 5.59E-05 & 7.43E-03 & 1.07E-02 \\
& AFNO-AR & 1.89E-05 & 1.45E-02 & 4.86E-02 \\
& \cellcolor{cyan!15}\textbf{AFNO} & \cellcolor{cyan!15}\textbf{2.50E-06} & \cellcolor{cyan!15}\textbf{1.57E-03} & \cellcolor{cyan!15}\textbf{1.00E-02} \\
\bottomrule
\end{tabular}
\end{table}

Table~\ref{tab:abtion} reports the rollout errors of all variants, where we present the results at 80 rollout steps for 1D-Burgers and at 20 rollout steps for 2D-NS. The complete AFNO consistently delivers the best performance across all datasets and evaluation metrics. In contrast, simplifying or replacing the latent flow matching module results in noticeable degradation in long-horizon prediction accuracy, with the degradation being particularly evident in the Rel.$H_1$. This behavior suggests that faithful modeling of continuous latent dynamics is critical for maintaining gradient consistency and long-horizon stability. Furthermore, replacing the encoder and decoder with purely convolutional residual architectures also leads to inferior performance, underscoring the importance of the FNO-based design in capturing global spatial structures and enabling stable extrapolation.

\begin{table}[h]
\centering
\caption{Sensitivity of AFNO to $C_z$ and $k$ on the 1D-Burgers and 2D-NS equations}
\label{tab:lfo_capacity_results}
\setlength{\tabcolsep}{4pt}
\renewcommand{\arraystretch}{1.05}
\footnotesize
\begin{tabular}{llccccc}
\toprule
Equation & Variant & $C_z$ & $k$ & MSE & Rel.$L_2$ & Rel.$H_1$ \\
\midrule
\multirow{2}{*}{1D-Burgers}
& AFNO-Small & 8  & 2 & 5.62E-14 & 3.11E-02 & 3.58E-01 \\
& AFNO-Large & 32 & 8 & 1.48E-14 & 1.71E-02 & 1.85E-01 \\
& \textbf{AFNO} & \textbf{16} & \textbf{4} & \textbf{1.33E-14}  & \textbf{1.60E-02} & \textbf{1.71E-01} \\
\midrule
\multirow{2}{*}{2D-NS}
& AFNO-Small & 8  & 2 & 8.95E-06 & 9.45E-03 & 2.84E-02 \\
& AFNO-Large & 32 & 8 & 2.85E-06 & 1.72E-03 & 1.15E-02 \\
& \textbf{AFNO} & \textbf{16} & \textbf{4} &\textbf{2.50E-06} & \textbf{1.57E-03} & \textbf{1.00E-02} \\
\bottomrule
\end{tabular}
\end{table}

\subsection{Parameter Sensitivity Analysis}

\noindent\textbf{Loss Weight Sensitivity.}
The training strategy adopts a two-stage weighting scheme, where the second stage jointly incorporates $\mathcal{L}_{\mathrm{rec}}$, $\mathcal{L}_{\mathrm{flow}}$, and $\mathcal{L}_{\mathrm{roll}}$. Based on this setting, three additional loss weight configurations are introduced in the second stage, corresponding to balanced weighting, flow-dominated training, and rollout-dominated training. Table~\ref{tab:sensitivity} reports the multi-step rollout errors under different weight configurations on the 1D-Burgers and 2D-NS equations. The results indicate that the overall performance of AFNO remains relatively stable across different loss weight allocations, with similar accuracy observed across the evaluation metrics, suggesting a certain degree of robustness to the choice of loss weights. In addition, moderately incorporating flow and rollout constraints contributes to improved Rel.$H_1$ performance, reflecting better preservation of gradient-level structures. The motivation for adopting a two-stage training strategy lies in balancing optimization stability and long-horizon dynamical modeling. During the initial stage, training with only the reconstruction loss allows the encoder and decoder to first learn stable and meaningful latent representations, thereby avoiding unstable gradient interference when the latent space is not yet well-formed. Building upon this foundation, the second stage gradually introduces the latent flow matching and rollout losses, which reduce optimization difficulty while maintaining stable convergence.

\begin{figure*}[t]
  \centering
  \includegraphics[width=\linewidth]{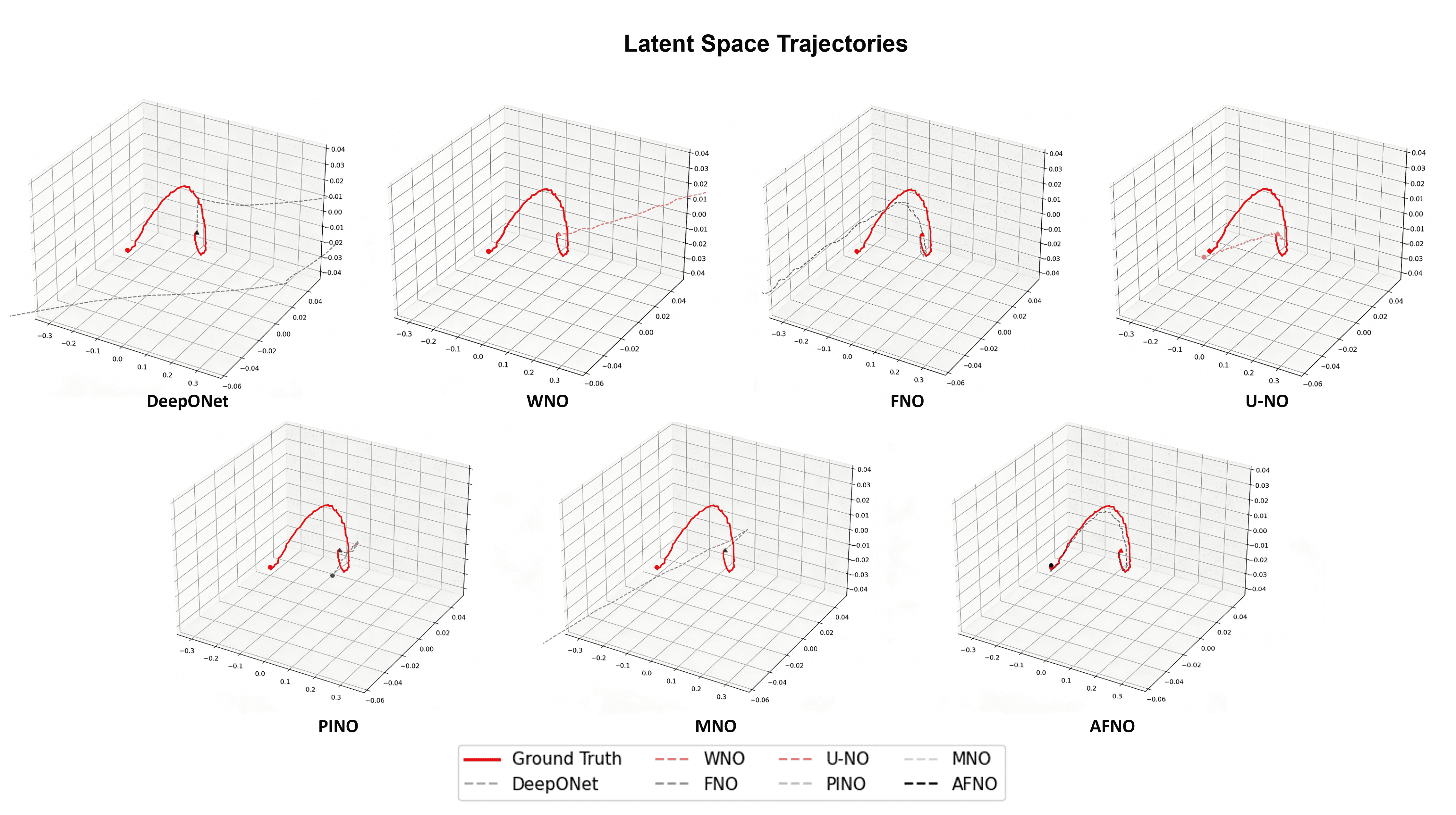}
   \caption{Latent space trajectory visualization for the 1D-Burgers equation.}
  \label{fig:6}
\end{figure*}

\noindent\textbf{Latent Capacity Sensitivity.}
We further examine the sensitivity of AFNO to the latent dimension and Fourier bandwidth. Specifically, we consider two capacity variants, \textbf{AFNO-Small} and \textbf{AFNO-Large}, which set the latent dimension and Fourier bandwidth to $(C_z,k)=(8,2)$ and $(32,8)$, respectively. These variants are introduced to examine whether the performance of AFNO is primarily associated with the proposed latent evolution design or with increased latent space capacity. Table~\ref{tab:lfo_capacity_results} summarizes the corresponding results on the 1D-Burgers and 2D-NS equations. The results show that reducing latent space capacity leads to a consistent decline in long-horizon prediction accuracy, whereas increasing the latent dimension and Fourier bandwidth results in only limited changes relative to the default setting. This suggests that the default configuration provides a suitable balance between representation capacity and dynamical modeling.

\begin{figure}[t]
  \centering
  \includegraphics[width=\linewidth]{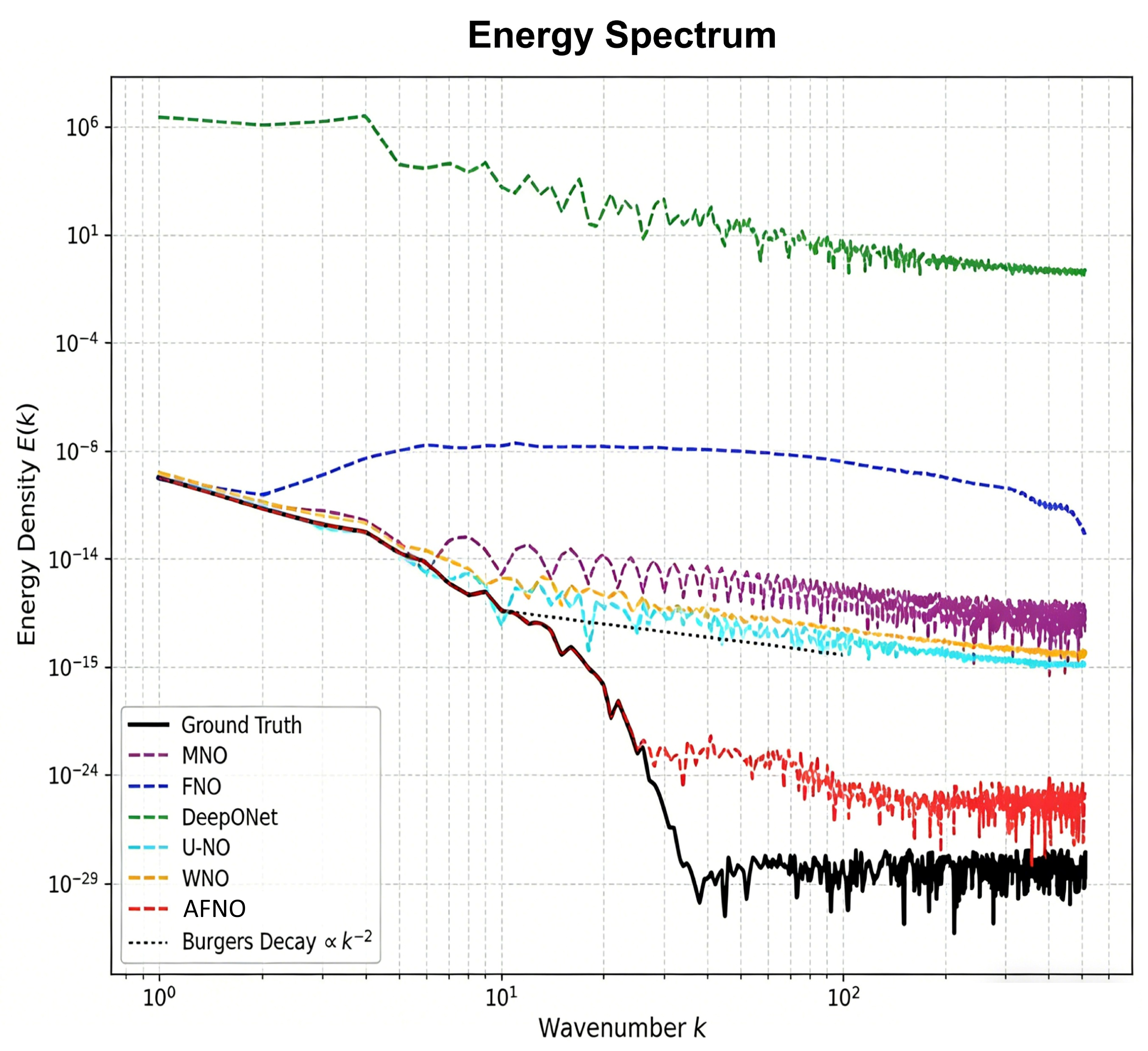}
   \caption{Energy spectrum comparison on the 1D-Burgers equation.}
  \label{fig:7}
\end{figure}

\subsection{Latent Trajectory Alignment and Spectral Consistency}
\begin{figure*}[h]
  \centering
  \includegraphics[width=\linewidth]{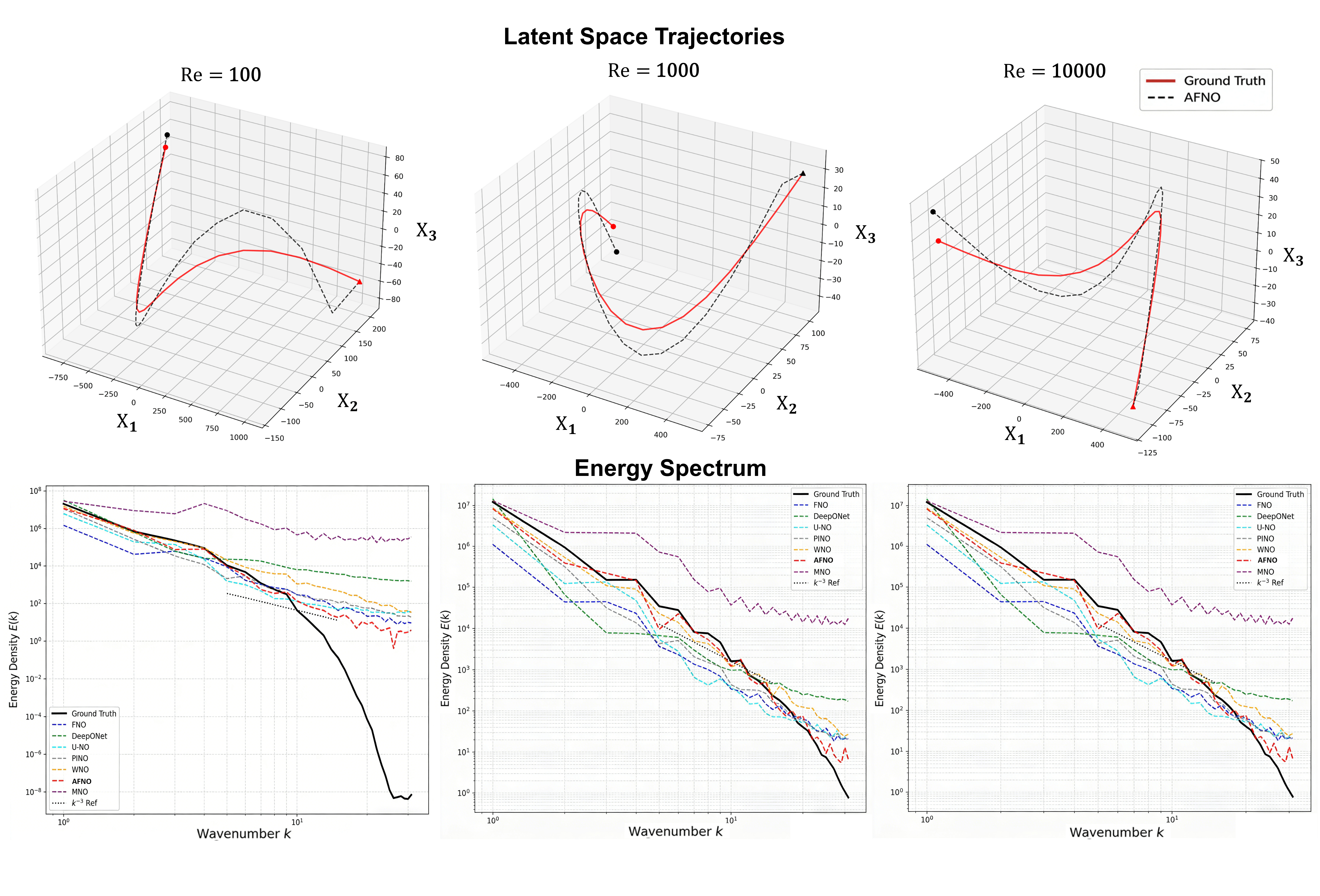}
   \caption{Latent trajectory (top) and energy spectrum visualization (bottom) for the 2D-NS equations under different parameters.}
  \label{fig:8}
\end{figure*}

Existing evaluation metrics primarily reflect pointwise numerical approximation accuracy, but they do not sufficiently characterize whether a model captures the underlying dynamical structure and spectral consistency of physical systems. To further analyze the dynamical information encoded in the learned representations, we visualize the latent trajectories induced by the model. Specifically, we apply PCA \cite{shlens2014tutorial} to project the high-dimensional latent space into a low-dimensional subspace and examine the evolution trajectories learned by AFNO for PDE prediction. Fig.~\ref{fig:6} compares the predicted and ground-truth trajectories for the 1D-Burgers equation, while the top panel of Fig.~\ref{fig:8} reports AFNO predictions on the parametric 2D-NS equations. These results indicate that AFNO maintains coherent latent trajectories and that the predicted trajectories remain closely aligned with the underlying manifold of the true physical system.

The bottom panels of Figs.~\ref{fig:8} and~\ref{fig:7} further compare AFNO with baseline methods in the frequency domain. 
For fluid systems with turbulence or shocks, such as the 1D-Burgers and 2D-NS equations, multi-scale energy cascades are essential, as energy is transferred from large to small scales through nonlinear interactions before viscous dissipation. 
Neural predictors often suffer from high-frequency spectral degradation, which weakens their ability to preserve fine-scale dynamics. 
Spectral comparisons show that AFNO better follows ground-truth spectral decay and retains meaningful energy across a broader range of wavenumbers, whereas baseline methods exhibit stronger high-frequency attenuation. 
This behavior is consistent with the flow formulation of AFNO.

\begin{table}[h]
\centering
\caption{Computational complexity and runtime comparison on the 1D-Burgers and 2D-NS equations, including single-step FLOPs, average training time per epoch, and wall-clock latency for 100-step rollout inference.}
\label{tab:complexity}
\setlength{\tabcolsep}{3.5pt}
\footnotesize
\begin{tabular}{lccc}
\toprule
Model & FLOPs (1-step) & Epoch time (s) & 100-step inference time (s) \\
\midrule

\multicolumn{4}{c}{\textbf{1D-Burgers}} \\
\midrule
FNO      & 8.92E+06 & 1.21E+00 & 1.57E-01 \\
DeepONet & 1.31E+08 & 1.78E+00 & 4.37E-02 \\
U-NO     & 7.86E+06 & 7.05E+00 & 1.70E-01 \\
PINO     & 8.92E+06 & 1.08E+00 & 1.56E-01 \\
WNO      & 1.99E+07 & 1.24E+00 & 9.38E-02 \\
MNO      & 3.10E+06 & 1.60E+00 & 9.38E-01 \\
AFNO     & 1.61E+07 & 1.15E+00 & 5.11E-01 \\

\midrule
\multicolumn{4}{c}{\textbf{2D-NS}} \\
\midrule
FNO      & 2.80E+09 & 2.01E+01 & 1.67E-01 \\
DeepONet & 1.99E+10 & 3.89E+00 & 5.14E-02 \\
U-NO     & 2.53E+09 & 2.45E+01 & 2.29E-01 \\
PINO     & 8.16E+08 & 7.63E+00 & 1.52E-01 \\
WNO      & 2.58E+10 & 8.33E+01 & 1.28E-01 \\
MNO      & 7.93E+07 & 1.26E+01 & 1.49E+00 \\
AFNO     & 4.79E+09 & 7.32E+00 & 8.17E-01 \\
\bottomrule
\end{tabular}
\end{table}

\subsection{Computational Complexity Analysis}
\label{sec:5.8}
To characterize the computational overhead associated with latent space rollout, Table~\ref{tab:complexity} compares AFNO with representative baselines in terms of single-step FLOPs, average training time per epoch, and wall-clock latency for 100-step rollout inference. Since AFNO performs explicit latent-state evolution during rollout, its long-horizon inference latency is higher than that of purely discrete one-step operator baselines such as FNO and PINO. The additional overhead, however, remains moderate. On 1D-Burgers, AFNO requires 0.511\,s for 100-step inference, which is lower than MNO (0.938\,s) and remains within the same order of magnitude as the other neural operator baselines. On 2D-NS, AFNO requires 0.817\,s for 100-step rollout, again lower than MNO (1.49\,s). In terms of training efficiency, AFNO achieves lower per-epoch training time than several baselines. Although AFNO is not the most efficient method, its computational overhead remains moderate relative to its gains in long-horizon stability.

\section{Conclusion}
In this paper, we propose Autoregression-Free Neural Operators (AFNO) for modeling the underlying PDE evolution and forecasting long-horizon rollouts. Instead of performing autoregressive rollout in the physical space, AFNO models temporal evolution in a parameter-conditioned latent space, where high-dimensional PDE states are reorganized into a compact and structured representation and the corresponding latent vector field is learned from discrete observations via flow matching. This formulation separates temporal evolution from recursive physical space prediction and supports generalization across different physical parameter configurations within the same PDE family. Extensive experiments on multiple canonical time-dependent PDEs show that AFNO consistently improves long-horizon prediction accuracy and stability, particularly in multi-step rollout and cross-parameter evaluation settings, where error accumulation is substantially reduced. These results indicate that autoregression-free latent space modeling provides an effective alternative to conventional autoregressive formulations for PDE learning. Future work will investigate extensions to stochastic dynamics, multiphysics systems, and adaptive latent representations, as well as theoretical connections between learned latent dynamics, invariant manifolds, and attractor geometry.


\bibliography{LFO}

@article{huang2025partial,
  title={Partial differential equations meet deep neural networks: A survey},
  author={Huang, Shudong and Feng, Wentao and Tang, Chenwei and He, Zhenan and Yu, Caiyang and Lv, Jiancheng},
  journal={IEEE Transactions on Neural Networks and Learning Systems},
  year={2025},
  publisher={IEEE}
}

@article{sharma2023review,
  title={A review of physics-informed machine learning in fluid mechanics},
  author={Sharma, Pushan and Chung, Wai Tong and Akoush, Bassem and Ihme, Matthias},
  journal={Energies},
  volume={16},
  number={5},
  pages={2343},
  year={2023},
  publisher={MDPI}
}

@article{mengaldo2019current,
  title={Current and emerging time-integration strategies in global numerical weather and climate prediction},
  author={Mengaldo, Gianmarco and Wyszogrodzki, Andrzej and Diamantakis, Michail and Lock, Sarah-Jane and Giraldo, Francis X and Wedi, Nils P},
  journal={Archives of Computational Methods in Engineering},
  volume={26},
  number={3},
  pages={663--684},
  year={2019},
  publisher={Springer}
}

@article{patawari2025traditional,
  title={From traditional to computationally efficient scientific computing algorithms in option pricing: Current progresses with future directions},
  author={Patawari, Arihant and Das, Pratibhamoy},
  journal={Archives of Computational Methods in Engineering},
  pages={1--40},
  year={2025},
  publisher={Springer}
}

@article{azizzadenesheli2024neural,
  title={Neural operators for accelerating scientific simulations and design},
  author={Azizzadenesheli, Kamyar and Kovachki, Nikola and Li, Zongyi and Liu-Schiaffini, Miguel and Kossaifi, Jean and Anandkumar, Anima},
  journal={Nature Reviews Physics},
  volume={6},
  number={5},
  pages={320--328},
  year={2024},
  publisher={Nature Publishing Group UK London}
}

@article{meng2025physics,
  title={When physics meets machine learning: A survey of physics-informed machine learning},
  author={Meng, Chuizheng and Griesemer, Sam and Cao, Defu and Seo, Sungyong and Liu, Yan},
  journal={Machine Learning for Computational Science and Engineering},
  volume={1},
  number={1},
  pages={20},
  year={2025},
  publisher={Springer}
}

@article{li2020fourier,
  title={Fourier neural operator for parametric partial differential equations},
  author={Li, Zongyi and Kovachki, Nikola and Azizzadenesheli, Kamyar and Liu, Burigede and Bhattacharya, Kaushik and Stuart, Andrew and Anandkumar, Anima},
  journal={arXiv preprint arXiv:2010.08895},
  year={2020}
}

@article{tadmor2012review,
  title={A review of numerical methods for nonlinear partial differential equations},
  author={Tadmor, Eitan},
  journal={Bulletin of the American Mathematical Society},
  volume={49},
  number={4},
  pages={507--554},
  year={2012}
}

@article{song2024seismic,
  title={Seismic traveltime simulation for variable velocity models using physics-informed Fourier neural operator},
  author={Song, Chao and Zhao, Tianshuo and Waheed, Umair Bin and Liu, Cai and Tian, You},
  journal={IEEE Transactions on Geoscience and Remote Sensing},
  volume={62},
  pages={1--9},
  year={2024},
  publisher={IEEE}
}

@article{liu2023domain,
  title={Domain agnostic fourier neural operators},
  author={Liu, Ning and Jafarzadeh, Siavash and Yu, Yue},
  journal={Advances in Neural Information Processing Systems},
  volume={36},
  pages={47438--47450},
  year={2023}
}

@article{you2025mscalefno,
  title={MscaleFNO: Multi-scale Fourier neural operator learning for oscillatory functions and wave scattering problems},
  author={You, Zilin and Xu, Zhenli and Cai, Wei},
  journal={Journal of Computational Physics},
  pages={114530},
  year={2025},
  publisher={Elsevier}
}

@article{kovachki2023neural,
  title={Neural operator: Learning maps between function spaces with applications to pdes},
  author={Kovachki, Nikola and Li, Zongyi and Liu, Burigede and Azizzadenesheli, Kamyar and Bhattacharya, Kaushik and Stuart, Andrew and Anandkumar, Anima},
  journal={Journal of Machine Learning Research},
  volume={24},
  number={89},
  pages={1--97},
  year={2023}
}

@article{tanyu2023deep,
  title={Deep learning methods for partial differential equations and related parameter identification problems},
  author={Tanyu, Derick Nganyu and Ning, Jianfeng and Freudenberg, Tom and Heilenk{\"o}tter, Nick and Rademacher, Andreas and Iben, Uwe and Maass, Peter},
  journal={Inverse Problems},
  volume={39},
  number={10},
  pages={103001},
  year={2023},
  publisher={IOP Publishing}
}

@inproceedings{wang2020towards,
  title={Towards physics-informed deep learning for turbulent flow prediction},
  author={Wang, Rui and Kashinath, Karthik and Mustafa, Mustafa and Albert, Adrian and Yu, Rose},
  booktitle={Proceedings of the 26th ACM SIGKDD international conference on knowledge discovery \& data mining},
  pages={1457--1466},
  year={2020}
}

@article{gupta2022three,
  title={Three-dimensional deep learning-based reduced order model for unsteady flow dynamics with variable Reynolds number},
  author={Gupta, Rachit and Jaiman, Rajeev},
  journal={Physics of Fluids},
  volume={34},
  number={3},
  year={2022},
  publisher={AIP Publishing}
}

@inproceedings{sun2020neupde,
  title={NeuPDE: Neural network based ordinary and partial differential equations for modeling time-dependent data},
  author={Sun, Yifan and Zhang, Linan and Schaeffer, Hayden},
  booktitle={Mathematical and Scientific Machine Learning},
  pages={352--372},
  year={2020},
  organization={PMLR}
}

@article{croitoru2023diffusion,
  title={Diffusion models in vision: A survey},
  author={Croitoru, Florinel-Alin and Hondru, Vlad and Ionescu, Radu Tudor and Shah, Mubarak},
  journal={IEEE transactions on pattern analysis and machine intelligence},
  volume={45},
  number={9},
  pages={10850--10869},
  year={2023},
  publisher={Ieee}
}

@article{zhao2024flowturbo,
  title={Flowturbo: Towards real-time flow-based image generation with velocity refiner},
  author={Zhao, Wenliang and Shi, Minglei and Yu, Xumin and Zhou, Jie and Lu, Jiwen},
  journal={Advances in Neural Information Processing Systems},
  volume={37},
  pages={4148--4176},
  year={2024}
}

@article{takamoto2022pdebench,
  title={Pdebench: An extensive benchmark for scientific machine learning},
  author={Takamoto, Makoto and Praditia, Timothy and Leiteritz, Raphael and MacKinlay, Daniel and Alesiani, Francesco and Pfl{\"u}ger, Dirk and Niepert, Mathias},
  journal={Advances in Neural Information Processing Systems},
  volume={35},
  pages={1596--1611},
  year={2022}
}

@article{elasri2022image,
  title={Image generation: A review},
  author={Elasri, Mohamed and Elharrouss, Omar and Al-Maadeed, Somaya and Tairi, Hamid},
  journal={Neural Processing Letters},
  volume={54},
  number={5},
  pages={4609--4646},
  year={2022},
  publisher={Springer}
}

@article{lu2019deeponet,
  title={Deeponet: Learning nonlinear operators for identifying differential equations based on the universal approximation theorem of operators},
  author={Lu, Lu and Jin, Pengzhan and Karniadakis, George Em},
  journal={arXiv preprint arXiv:1910.03193},
  year={2019}
}

@article{li2020neural,
  title={Neural operator: Graph kernel network for partial differential equations},
  author={Li, Zongyi and Kovachki, Nikola and Azizzadenesheli, Kamyar and Liu, Burigede and Bhattacharya, Kaushik and Stuart, Andrew and Anandkumar, Anima},
  journal={arXiv preprint arXiv:2003.03485},
  year={2020}
}

@article{li2023fourier,
  title={Fourier neural operator with learned deformations for pdes on general geometries},
  author={Li, Zongyi and Huang, Daniel Zhengyu and Liu, Burigede and Anandkumar, Anima},
  journal={Journal of Machine Learning Research},
  volume={24},
  number={388},
  pages={1--26},
  year={2023}
}

@article{li2024physics,
  title={Physics-informed neural operator for learning partial differential equations},
  author={Li, Zongyi and Zheng, Hongkai and Kovachki, Nikola and Jin, David and Chen, Haoxuan and Liu, Burigede and Azizzadenesheli, Kamyar and Anandkumar, Anima},
  journal={ACM/IMS Journal of Data Science},
  volume={1},
  number={3},
  pages={1--27},
  year={2024},
  publisher={ACM New York, NY}
}

@article{tripura2023wavelet,
  title={Wavelet neural operator for solving parametric partial differential equations in computational mechanics problems},
  author={Tripura, Tapas and Chakraborty, Souvik},
  journal={Computer Methods in Applied Mechanics and Engineering},
  volume={404},
  pages={115783},
  year={2023},
  publisher={Elsevier}
}

@article{raonic2023convolutional,
  title={Convolutional neural operators for robust and accurate learning of PDEs},
  author={Raonic, Bogdan and Molinaro, Roberto and De Ryck, Tim and Rohner, Tobias and Bartolucci, Francesca and Alaifari, Rima and Mishra, Siddhartha and de B{\'e}zenac, Emmanuel},
  journal={Advances in Neural Information Processing Systems},
  volume={36},
  pages={77187--77200},
  year={2023}
}

@inproceedings{hao2023gnot,
  title={Gnot: A general neural operator transformer for operator learning},
  author={Hao, Zhongkai and Wang, Zhengyi and Su, Hang and Ying, Chengyang and Dong, Yinpeng and Liu, Songming and Cheng, Ze and Song, Jian and Zhu, Jun},
  booktitle={International Conference on Machine Learning},
  pages={12556--12569},
  year={2023},
  organization={PMLR}
}

@article{vaswani2017attention,
  title={Attention is all you need},
  author={Vaswani, Ashish and Shazeer, Noam and Parmar, Niki and Uszkoreit, Jakob and Jones, Llion and Gomez, Aidan N and Kaiser, {\L}ukasz and Polosukhin, Illia},
  journal={Advances in neural information processing systems},
  volume={30},
  year={2017}
}

@inproceedings{gu2024mamba,
  title={Mamba: Linear-time sequence modeling with selective state spaces},
  author={Gu, Albert and Dao, Tri},
  booktitle={First conference on language modeling},
  year={2024}
}

@article{cheng2025mamba,
  title={Mamba neural operator: Who wins? transformers vs. state-space models for pdes},
  author={Cheng, Chun-Wun and Huang, Jiahao and Zhang, Yi and Yang, Guang and Sch{\"o}nlieb, Carola-Bibiane and Aviles-Rivero, Angelica I},
  journal={Journal of Computational Physics},
  pages={114567},
  year={2025},
  publisher={Elsevier}
}

@inproceedings{duan2024weditgan,
  title={Weditgan: Few-shot image generation via latent space relocation},
  author={Duan, Yuxuan and Niu, Li and Hong, Yan and Zhang, Liqing},
  booktitle={Proceedings of the AAAI conference on artificial intelligence},
  volume={38},
  number={2},
  pages={1653--1661},
  year={2024}
}

@inproceedings{gelada2019deepmdp,
  title={Deepmdp: Learning continuous latent space models for representation learning},
  author={Gelada, Carles and Kumar, Saurabh and Buckman, Jacob and Nachum, Ofir and Bellemare, Marc G},
  booktitle={International conference on machine learning},
  pages={2170--2179},
  year={2019},
  organization={PMLR}
}

@article{kingma2013auto,
  title={Auto-encoding variational bayes},
  author={Kingma, Diederik P and Welling, Max},
  journal={arXiv preprint arXiv:1312.6114},
  year={2013}
}

@article{goodfellow2020generative,
  title={Generative adversarial networks},
  author={Goodfellow, Ian and Pouget-Abadie, Jean and Mirza, Mehdi and Xu, Bing and Warde-Farley, David and Ozair, Sherjil and Courville, Aaron and Bengio, Yoshua},
  journal={Communications of the ACM},
  volume={63},
  number={11},
  pages={139--144},
  year={2020},
  publisher={ACM New York, NY, USA}
}

@article{dinh2016density,
  title={Density estimation using real nvp},
  author={Dinh, Laurent and Sohl-Dickstein, Jascha and Bengio, Samy},
  journal={arXiv preprint arXiv:1605.08803},
  year={2016}
}

@article{kingma2018glow,
  title={Glow: Generative flow with invertible 1x1 convolutions},
  author={Kingma, Durk P and Dhariwal, Prafulla},
  journal={Advances in neural information processing systems},
  volume={31},
  year={2018}
}

@article{ho2020denoising,
  title={Denoising diffusion probabilistic models},
  author={Ho, Jonathan and Jain, Ajay and Abbeel, Pieter},
  journal={Advances in neural information processing systems},
  volume={33},
  pages={6840--6851},
  year={2020}
}

@inproceedings{devlin2019bert,
  title={Bert: Pre-training of deep bidirectional transformers for language understanding},
  author={Devlin, Jacob and Chang, Ming-Wei and Lee, Kenton and Toutanova, Kristina},
  booktitle={Proceedings of the 2019 conference of the North American chapter of the association for computational linguistics: human language technologies, volume 1 (long and short papers)},
  pages={4171--4186},
  year={2019}
}

@article{van2017neural,
  title={Neural discrete representation learning},
  author={Van Den Oord, Aaron and Vinyals, Oriol and others},
  journal={Advances in neural information processing systems},
  volume={30},
  year={2017}
}

@article{van2016wavenet,
  title={Wavenet: A generative model for raw audio},
  author={Dieleman, Sander and Zen, Heiga and Simonyan, Karen and Vinyals, Oriol and Graves, Alex and Kalchbrenner, Nal and Senior, Andrew and Kavukcuoglu, Koray and others},
  journal={arXiv preprint arXiv:1609.03499},
  volume={12},
  pages={1},
  year={2016}
}

@article{wang2024latent,
  title={Latent neural operator for solving forward and inverse pde problems},
  author={Wang, Tian and Wang, Chuang},
  journal={Advances in Neural Information Processing Systems},
  volume={37},
  pages={33085--33107},
  year={2024}
}

@article{hou2007computing,
  title={Computing nearly singular solutions using pseudo-spectral methods},
  author={Hou, Thomas Y and Li, Ruo},
  journal={Journal of Computational Physics},
  volume={226},
  number={1},
  pages={379--397},
  year={2007},
  publisher={Elsevier}
}

@inproceedings{crank1947practical,
  title={A practical method for numerical evaluation of solutions of partial differential equations of the heat-conduction type},
  author={Crank, John and Nicolson, Phyllis},
  booktitle={Mathematical proceedings of the Cambridge philosophical society},
  volume={43},
  number={1},
  pages={50--67},
  year={1947},
  organization={Cambridge University Press}
}

@article{rahman2022u,
  title={U-no: U-shaped neural operators},
  author={Rahman, Md Ashiqur and Ross, Zachary E and Azizzadenesheli, Kamyar},
  journal={arXiv preprint arXiv:2204.11127},
  year={2022}
}

@article{brandstetter2022message,
  title={Message passing neural PDE solvers},
  author={Brandstetter, Johannes and Worrall, Daniel and Welling, Max},
  journal={arXiv preprint arXiv:2202.03376},
  year={2022}
}

@article{kingma2014adam,
  title={Adam: A method for stochastic optimization},
  author={Kingma, Diederik P},
  journal={arXiv preprint arXiv:1412.6980},
  year={2014}
}

@inproceedings{he2016deep,
  title={Deep residual learning for image recognition},
  author={He, Kaiming and Zhang, Xiangyu and Ren, Shaoqing and Sun, Jian},
  booktitle={Proceedings of the IEEE conference on computer vision and pattern recognition},
  pages={770--778},
  year={2016}
}

@article{rumelhart1986learning,
  title={Learning representations by back-propagating errors},
  author={Rumelhart, David E and Hinton, Geoffrey E and Williams, Ronald J},
  journal={nature},
  volume={323},
  number={6088},
  pages={533--536},
  year={1986},
  publisher={Nature Publishing Group UK London}
}

@article{ye2025recurrent,
  title={Recurrent Neural Operators: Stable Long-Term PDE Prediction},
  author={Ye, Zaijun and Zhang, Chen-Song and Wang, Wansheng},
  journal={arXiv preprint arXiv:2505.20721},
  year={2025}
}

@inproceedings{ahn2025lightweight,
  title={Lightweight Fourier Neural Operator for Time-Dependent Partial Differential Equations},
  author={Dawon Ahn and Satish Chandran and Daniel Leibovici and Nikola Kovachki and Vagelis Papalexakis and Jean Kossaifi},
  booktitle={Proceedings of the 39th Conference on Neural Information Processing Systems (NeurIPS 2025) - Workshop on Machine Learning and the Physical Sciences},
  year={2025},
  url={https://neurips.cc/virtual/2025/loc/san-diego/123080}
}

@article{chen2025freqmoe,
  title={FreqMoE: Dynamic Frequency Enhancement for Neural PDE Solvers},
  author={Chen, Tianyu and Zhou, Haoyi and Li, Ying and Wang, Hao and Zhang, Zhenzhe and Zhu, Tianchen and Zhang, Shanghang and Li, Jianxin},
  journal={arXiv preprint arXiv:2505.06858},
  year={2025}
}

@article{li2026sgno,
  title={SGNO: Spectral Generator Neural Operators for Stable Long Horizon PDE Rollouts},
  author={Li, Jiayi and Wang, Zhaonan and Salim, Flora D},
  journal={arXiv preprint arXiv:2602.18801},
  year={2026}
}

@article{dao2023flow,
  title={Flow matching in latent space},
  author={Dao, Quan and Phung, Hao and Nguyen, Binh and Tran, Anh},
  journal={arXiv preprint arXiv:2307.08698},
  year={2023}
}

@incollection{nussbaumer1981fast,
  title={The fast Fourier transform},
  author={Nussbaumer, Henri J},
  booktitle={Fast Fourier transform and convolution algorithms},
  pages={80--111},
  year={1981},
  publisher={Springer}
}

@article{zhou2023physics,
  title={Physics-informed neural networks for solving time-dependent mode-resolved phonon Boltzmann transport equation},
  author={Zhou, Jiahang and Li, Ruiyang and Luo, Tengfei},
  journal={npj Computational Materials},
  volume={9},
  number={1},
  pages={212},
  year={2023},
  publisher={Nature Publishing Group UK London}
}

@article{diab2025temporal,
  title={Temporal neural operator for modeling time-dependent physical phenomena},
  author={Diab, Waleed and Al Kobaisi, Mohammed},
  journal={Scientific Reports},
  volume={15},
  number={1},
  pages={32791},
  year={2025},
  publisher={Nature Publishing Group UK London}
}

@article{oommen2024rethinking,
  title={Rethinking materials simulations: Blending direct numerical simulations with neural operators},
  author={Oommen, Vivek and Shukla, Khemraj and Desai, Saaketh and Dingreville, R{\'e}mi and Karniadakis, George Em},
  journal={npj Computational Materials},
  volume={10},
  number={1},
  pages={145},
  year={2024},
  publisher={Nature Publishing Group UK London}
}

@inproceedings{rombach2022high,
  title={High-resolution image synthesis with latent diffusion models},
  author={Rombach, Robin and Blattmann, Andreas and Lorenz, Dominik and Esser, Patrick and Ommer, Bj{\"o}rn},
  booktitle={Proceedings of the IEEE/CVF conference on computer vision and pattern recognition},
  pages={10684--10695},
  year={2022}
}

@article{cuomo2022scientific,
  title={Scientific machine learning through physics--informed neural networks: Where we are and what’s next},
  author={Cuomo, Salvatore and Di Cola, Vincenzo Schiano and Giampaolo, Fabio and Rozza, Gianluigi and Raissi, Maziar and Piccialli, Francesco},
  journal={Journal of Scientific Computing},
  volume={92},
  number={3},
  pages={88},
  year={2022},
  publisher={Springer}
}

@article{bengio2015scheduled,
  title={Scheduled sampling for sequence prediction with recurrent neural networks},
  author={Bengio, Samy and Vinyals, Oriol and Jaitly, Navdeep and Shazeer, Noam},
  journal={Advances in neural information processing systems},
  volume={28},
  year={2015}
}

@article{chen2018neural,
  title={Neural ordinary differential equations},
  author={Chen, Ricky TQ and Rubanova, Yulia and Bettencourt, Jesse and Duvenaud, David K},
  journal={Advances in neural information processing systems},
  volume={31},
  year={2018}
}

@article{rubanova2019latent,
  title={Latent ordinary differential equations for irregularly-sampled time series},
  author={Rubanova, Yulia and Chen, Ricky TQ and Duvenaud, David K},
  journal={Advances in neural information processing systems},
  volume={32},
  year={2019}
}

@inproceedings{ross2011reduction,
  title={A reduction of imitation learning and structured prediction to no-regret online learning},
  author={Ross, St{\'e}phane and Gordon, Geoffrey and Bagnell, Drew},
  booktitle={Proceedings of the fourteenth international conference on artificial intelligence and statistics},
  pages={627--635},
  year={2011},
  organization={JMLR Workshop and Conference Proceedings}
}

@article{shlens2014tutorial,
  title={A tutorial on principal component analysis},
  author={Shlens, Jonathon},
  journal={arXiv preprint arXiv:1404.1100},
  year={2014}
}
\bibliographystyle{IEEEtran}


 




\vfill

\newpage

{\appendices

\section{Benchmark Designs}
This section provides additional details on the specific design of the PDEs used in the experiments.
\label{app:1.1}
This section describes the specific 1D and 2D PDEs used in the experiments. All equations are implemented based on the standardized data generation procedures of PDEBench, with appropriate extensions to provide a comprehensive evaluation of AFNO. The configurations are as follows:

\textbf{1D Burgers equation.}  
The Burgers equation is a classical model for studying the interaction between nonlinear advection and viscous dissipation. Its dynamics are governed by the nonlinear convection term $u u_x$ and the viscous diffusion term $\nu u_{xx}$, which produce steep gradients that smooth over time under viscous effects. The governing equation is
\begin{equation}
\frac{\partial u}{\partial t} + u \frac{\partial u}{\partial x} = \nu \frac{\partial^2 u}{\partial x^2}, \quad x \in [0, 2\pi).
\end{equation}
The viscosity coefficient is set to $\nu=0.01$. Spatial discretization uses a pseudospectral method, the linear diffusion term is integrated via the Crank-Nicolson scheme, and the nonlinear convection term is advanced using the Adams-Bashforth method. The spatial resolution is $N_x=1024$, and the temporal domain is $t\in[0,1.0]$, uniformly discretized into 101 points including the initial time.

\textbf{1D Allen-Cahn equation} is a prototypical reaction-diffusion system originating from mean-field theory for alloy phase separation. It describes the evolution of phase separation and interface dynamics under a double-well potential $F(u)=(u^2-1)^2/4$, where the solution typically splits into two stable phases $u\approx\pm 1$ with narrow transition layers. The governing equation is
\begin{equation}
\frac{\partial u}{\partial t} - \epsilon \frac{\partial^2 u}{\partial x^2} + \gamma (u^3 - u) = 0, \quad x \in [0,1.0].
\end{equation}
The interface width parameter is set to $\epsilon=10^{-4}$, and the reaction rate is $\gamma=5.0$. The solution is computed using a fourth-order integrating factor Runge--Kutta method (IFRK4) with an internal time step of approximately $10^{-4}$ to resolve the interface. The spatial resolution is $N_x=1024$, and the temporal domain is $t\in[0,1.0]$, uniformly discretized into 101 points.

\textbf{1D Kuramoto-Sivashinsky equation} is a canonical model for deterministic spatiotemporal chaos. It contains a negative diffusion term $u_{xx}$ and a fourth-order hyperdiffusion term $u_{xxxx}$, whose competition produces sustained chaotic evolution in the absence of external noise. The governing equation is
\begin{equation}
\frac{\partial u}{\partial t} + u \frac{\partial u}{\partial x} + \frac{\partial^2 u}{\partial x^2} + \frac{\partial^4 u}{\partial x^4} = 0, \quad x \in [0,64].
\end{equation}
The spatial domain length is $L=64.0$. Time integration uses an integrating factor method with 4 substeps for numerical stability. The spatial resolution is $N_x=1024$, and the temporal domain is $t\in[0,10.0]$, uniformly discretized into 101 points.

\textbf{2D Navier-Stokes equations} in vorticity-streamfunction form describes incompressible fluid dynamics, capturing energy cascades and vortex evolution. The governing equation is
\begin{equation}
\frac{\partial \omega}{\partial t} + (\mathbf{u}\cdot\nabla)\omega = \nu \Delta \omega + f, \quad \nabla\cdot\mathbf{u}=0.
\end{equation}
The Reynolds number is set to $Re=1000$ (corresponding to $\nu=10^{-3}$). Additional datasets at $Re=100$ and $Re=10000$ are generated to cover different flow regimes. Spatial derivatives are computed spectrally, and the linear viscous term is integrated using the Crank-Nicolson scheme. The computational domain has resolution $64\times64$, and the temporal domain is $t\in[0,20.0]$, sampled uniformly at 21 points (every 1.0 time unit).

\textbf{2D shallow water equations} describe the evolution of a free-surface fluid under gravity, capturing the nonlinear coupling between the fluid height $h$ and horizontal velocity field $\mathbf{u}=(u,v)$. The governing equations are
\begin{equation}
\begin{cases}
h_t + \nabla \cdot (h \mathbf{u}) = 0,\\
\mathbf{u}_t + (\mathbf{u} \cdot \nabla)\mathbf{u} + g \nabla h = 0.
\end{cases}
\end{equation}
The gravitational acceleration is set to $g=1.0$, and the mean water depth is $H_0=1.0$. Initial conditions are generated with random Gaussian perturbations. The right-hand side is constructed in rotational form, and time integration uses a fourth-order Runge-Kutta scheme with a 2/3 de-aliasing rule. The spatial resolution is $64\times64$, and the temporal domain is $t\in[0,2.0]$, uniformly discretized into 21 points.

\textbf{2D complex Ginzburg-Landau equation.}  
The complex Ginzburg--Landau equation is a general model for nonlinear oscillatory media and exhibits defect turbulence and spiral wave formation under certain parameter regimes. The governing equation is
\begin{equation}
\frac{\partial A}{\partial t} = A + (1 + ib)\Delta A - (1 + ic)|A|^2 A.
\end{equation}
The parameters are set to $b=2.0$ and $c=-2.0$. The solution is computed using a fourth-order integrating factor Runge--Kutta method with an internal time step of approximately $10^{-3}$. The spatial resolution is $128\times128$, and the temporal domain is $t\in[0,2.0]$, uniformly discretized into 21 points.

\section{Theoretical Analysis}
\label{app:2}

This section provides detailed proofs and extended analysis for the error propagation results discussed in the main paper. 
The main paper uses the autoregressive error bound to motivate the limitation of recursive physical-space rollout. 
We first give the detailed proof for an FNO-type autoregressive propagator as a representative baseline, and then extend the analysis to the latent-space continuous evolution adopted by AFNO.

\subsection{Autoregressive FNO-type rollout}
For time-dependent PDEs, a standard FNO-based predictor learns a one-step physical-space propagation operator 
$F:\mathcal{U}\rightarrow\mathcal{U}$ to approximate the exact solution operator $\Phi_{\Delta t}$, namely
\begin{equation}
F(u_t)\approx \Phi_{\Delta t}(u_t)=u_{t+\Delta t}.
\end{equation}
During long-horizon inference, the prediction is performed autoregressively:
\begin{equation}
\hat u_{t+\Delta t}=F(\hat u_t),
\end{equation}
where the predicted state $\hat u_t$ is recursively used as the input for the next step. 
This FNO-type rollout represents the standard physical-space autoregressive prediction paradigm analyzed in the main paper.

\noindent\textbf{\textit{Assumption 1 (Lipschitz Continuity of the Autoregressive Propagator).}} 
The learned one-step propagator $F$ is Lipschitz continuous on the physical state space $\mathcal{U}\subseteq\mathbb{R}^{d_u}$. 
That is, there exists a constant $L_F>0$ such that for any $u,u'\in\mathcal{U}$,
\begin{equation}
\|F(u)-F(u')\|\le L_F\|u-u'\|.
\end{equation}

\noindent\textbf{\textit{Proposition 1 (Detailed Proof of Autoregressive Error Propagation).}} 
Under Assumption 1, let $e(t)=\hat u_t-u_t$ denote the accumulated rollout error at time $t$. 
Define the local one-step approximation error as
\begin{equation}
\epsilon_t^F=\|F(u_t)-\Phi_{\Delta t}(u_t)\|.
\end{equation}
Then the single-step error propagation satisfies
\begin{equation}
\|e(t+\Delta t)\|\le L_F\|e(t)\|+\epsilon_t^F.
\end{equation}
After $n$ autoregressive rollout steps, the accumulated error is bounded by
\begin{equation}
\|e(t+n\Delta t)\|
\le
L_F^n\|e(t)\|
+
\sum_{i=0}^{n-1}L_F^i
\epsilon^F_{t+(n-1-i)\Delta t}.
\end{equation}
If $\epsilon_t^F\le \bar{\epsilon}_F$ uniformly, then
\begin{equation}
\|e(t+n\Delta t)\|
\le
L_F^n\|e(t)\|
+
\bar{\epsilon}_F\sum_{i=0}^{n-1}L_F^i.
\end{equation}

\noindent\textit{Proof.} 
By definition,
\begin{equation}
\begin{aligned}
e(t+\Delta t)
&=
\hat u_{t+\Delta t}-u_{t+\Delta t} \\
&=
F(\hat u_t)-\Phi_{\Delta t}(u_t).
\end{aligned}
\end{equation}
Adding and subtracting $F(u_t)$ gives
\begin{equation}
e(t+\Delta t)
=
\underbrace{F(\hat u_t)-F(u_t)}_{\text{input perturbation}}
+
\underbrace{F(u_t)-\Phi_{\Delta t}(u_t)}_{\text{local approximation error}}.
\end{equation}
Taking norms and applying Assumption 1 yields
\begin{equation}
\begin{aligned}
\|e(t+\Delta t)\|
&\le
\|F(\hat u_t)-F(u_t)\|
+
\|F(u_t)-\Phi_{\Delta t}(u_t)\|  \\
&\le
L_F\|\hat u_t-u_t\|+\epsilon_t^F \\
&=
L_F\|e(t)\|+\epsilon_t^F .
\end{aligned}
\end{equation}
Applying this recursion repeatedly gives
\begin{equation}
\|e(t+n\Delta t)\|
\le
L_F^n\|e(t)\|
+
\sum_{i=0}^{n-1}L_F^i
\epsilon^F_{t+(n-1-i)\Delta t}.
\end{equation}
When $\epsilon_t^F\le \bar{\epsilon}_F$, the uniform bound follows immediately. 
This completes the proof.

\subsection{Extended latent-space analysis for AFNO}
The preceding result explains the error amplification mechanism of autoregressive physical-space rollout. 
We now analyze AFNO, which replaces repeated physical-space propagation with continuous evolution in a structured latent space.

\noindent\textbf{\textit{Assumption 2 (Regularity of the Learned Latent Dynamics).}} 
Let the latent dynamics of AFNO be modeled by a parameter-conditioned vector field 
$v_\theta:\mathcal{Z}\times\mathcal{B}\rightarrow\mathcal{Z}$, where $\mathcal{Z}\subseteq\mathbb{R}^{d_z}$ denotes the latent space and $\mathcal{B}$ denotes the physical parameter space. 
For any fixed parameter embedding $b\in\mathcal{B}$, we assume that $v_\theta(\cdot,b)$ is Lipschitz continuous with constant $L_v>0$, i.e.,
\begin{equation}
\|v_\theta(z,b)-v_\theta(z',b)\|
\le
L_v\|z-z'\|,
\qquad
\forall z,z'\in\mathcal{Z}.
\end{equation}

\noindent\textbf{\textit{Assumption 3 (Lipschitz Continuity of the Decoder).}} 
Let the decoder $\mathrm{Dec}:\mathcal{Z}\rightarrow\mathcal{U}$ be Lipschitz continuous with constant $L_{\mathrm{dec}}>0$, such that
\begin{equation}
\|\mathrm{Dec}(z)-\mathrm{Dec}(z')\|
\le
L_{\mathrm{dec}}\|z-z'\|,
\qquad
\forall z,z'\in\mathcal{Z}.
\end{equation}

\noindent\textbf{\textit{Proposition 2 (Error Propagation of AFNO Latent Evolution).}} 
Under Assumptions 2 and 3, let the latent-space error be defined as
\begin{equation}
e_z(t)=\hat z_t-z_t.
\end{equation}
The predicted latent trajectory is updated by explicit Euler integration:
\begin{equation}
\hat z_{t+\Delta t}
=
\hat z_t+\Delta t\,v_\theta(\hat z_t,b),
\end{equation}
while the exact latent evolution is represented by the latent one-step flow
\begin{equation}
z_{t+\Delta t}=\Psi_{\Delta t}(z_t).
\end{equation}
Define the local latent consistency error as
\begin{equation}
\epsilon_t^z
=
\left\|
z_t+\Delta t\,v_\theta(z_t,b)-\Psi_{\Delta t}(z_t)
\right\|.
\end{equation}
Then the single-step latent error satisfies
\begin{equation}
\|e_z(t+\Delta t)\|
\le
(1+\Delta tL_v)\|e_z(t)\|+\epsilon_t^z.
\end{equation}
After $n$ extrapolation steps, the latent error is bounded by
\begin{equation}
\begin{aligned}
\|e_z(t+n\Delta t)\|
\le&
(1+\Delta tL_v)^n\|e_z(t)\| \\
&+
\sum_{i=0}^{n-1}
(1+\Delta tL_v)^i
\epsilon^z_{t+(n-1-i)\Delta t}.
\end{aligned}
\end{equation}
If $\epsilon_t^z\le \bar{\epsilon}_z$ uniformly, then
\begin{equation}
\|e_z(t+n\Delta t)\|
\le
(1+\Delta tL_v)^n\|e_z(t)\|
+
\bar{\epsilon}_z
\sum_{i=0}^{n-1}(1+\Delta tL_v)^i.
\end{equation}
The corresponding physical-space error satisfies
\begin{equation}
\|e_u(t+n\Delta t)\|
\le
L_{\mathrm{dec}}\|e_z(t+n\Delta t)\|.
\end{equation}

\noindent\textit{Proof.}
By definition,
\begin{equation}
\begin{aligned}
e_z(t+\Delta t)
&=
\hat z_{t+\Delta t}-z_{t+\Delta t} \\
&=
\hat z_t+\Delta t\,v_\theta(\hat z_t,b)-\Psi_{\Delta t}(z_t).
\end{aligned}
\end{equation}
Adding and subtracting $z_t+\Delta t\,v_\theta(z_t,b)$ gives
\begin{equation}
\begin{aligned}
e_z(t+\Delta t)
=&
\underbrace{
\hat z_t-z_t
+
\Delta t\left(v_\theta(\hat z_t,b)-v_\theta(z_t,b)\right)
}_{\text{latent error amplification}}
\\
&+
\underbrace{
z_t+\Delta t\,v_\theta(z_t,b)-\Psi_{\Delta t}(z_t)
}_{\text{local consistency error}} .
\end{aligned}
\end{equation}
Taking norms and applying Assumption 2 yields
\begin{equation}
\begin{aligned}
\|e_z(t+\Delta t)\|
&\le
\|e_z(t)\|
+
\Delta t L_v\|e_z(t)\|
+
\epsilon_t^z \\
&=
(1+\Delta tL_v)\|e_z(t)\|+\epsilon_t^z.
\end{aligned}
\end{equation}
Applying this recursion iteratively gives
\begin{equation}
\begin{aligned}
\|e_z(t+n\Delta t)\|
\le&
(1+\Delta tL_v)^n\|e_z(t)\| \\
&+
\sum_{i=0}^{n-1}
(1+\Delta tL_v)^i
\epsilon^z_{t+(n-1-i)\Delta t}.
\end{aligned}
\end{equation}
Finally, since
\begin{equation}
e_u(t)=\mathrm{Dec}(\hat z_t)-\mathrm{Dec}(z_t),
\end{equation}
Assumption 3 gives
\begin{equation}
\|e_u(t)\|
\le
L_{\mathrm{dec}}\|e_z(t)\|.
\end{equation}
This completes the proof.

\noindent\textbf{\textit{Physical-space error with reconstruction error.}}
The bound above assumes that the physical state can be exactly represented by the decoder. 
In practice, the encoder-decoder pair may introduce a reconstruction error. 
Let
\begin{equation}
\epsilon_{\mathrm{rec}}(t)=\|\mathrm{Dec}(z_t)-u_t\|.
\end{equation}
Then the physical-space prediction error can be decomposed as
\begin{equation}
\begin{aligned}
\|\hat u_t-u_t\|
&=
\|\mathrm{Dec}(\hat z_t)-u_t\| \\
&\le
\|\mathrm{Dec}(\hat z_t)-\mathrm{Dec}(z_t)\|
+
\|\mathrm{Dec}(z_t)-u_t\| \\
&\le
L_{\mathrm{dec}}\|\hat z_t-z_t\|
+
\epsilon_{\mathrm{rec}}(t).
\end{aligned}
\end{equation}
Thus, the final physical-space error consists of the latent rollout error amplified by the decoder and the reconstruction error of the latent representation.

\noindent\textbf{\textit{Analysis and Discussion.}}
The above derivations extend the main-paper analysis from autoregressive physical-space rollout to AFNO latent-space evolution. 
For an FNO-type autoregressive propagator, the homogeneous error factor is $L_F^n$, which reflects repeated application of the learned physical-space operator. 
For AFNO, the corresponding homogeneous factor is $(1+\Delta tL_v)^n$, while the accumulated error additionally depends on the local latent consistency errors $\epsilon_t^z$.

When $\Delta tL_v\ll 1$, using $\ln(1+x)\approx x$ for small $x$, we have
\begin{equation}
\begin{aligned}
(1+\Delta tL_v)^n
&=
\exp\!\left(n\ln(1+\Delta tL_v)\right) \\
&\approx
\exp(n\Delta tL_v)
=
\exp(L_vT),
\end{aligned}
\end{equation}
where $T=n\Delta t$ denotes the physical prediction horizon. 
This indicates that the homogeneous latent-space error growth is governed by the physical time horizon and the regularity of the learned vector field, rather than by recursive propagation in the high-dimensional physical field space.

The local consistency error $\epsilon_t^z$ measures the discrepancy between the learned Euler update and the exact latent one-step flow. 
It includes both the approximation error of the learned vector field and the discretization error induced by the numerical integrator. 
Moreover, the rollout loss $\mathcal{L}_{\mathrm{roll}}$ explicitly penalizes multi-step prediction deviations during training, thereby reducing the accumulated contribution of local consistency errors and improving long-horizon extrapolation stability.

\subsection{Structural comparison under fair one-step consistency}
The preceding bounds characterize the error propagation of autoregressive physical-space rollout and AFNO latent-space evolution separately.
We further compare their accumulated rollout behavior under a fair one-step consistency condition.
This comparison is intended to isolate the structural effect of error injection and propagation, rather than to assume a smaller one-step fitting error for AFNO.

\noindent\textbf{\textit{Assumption 4 (Fair one-step consistency).}}
Let $\eta>0$ denote a uniform bound on the teacher-forcing one-step prediction error under the same data, time step $\Delta t$, training objective, and convergence level.
We assume that the FNO-type autoregressive propagator and AFNO have comparable one-step physical-space errors on ground-truth inputs:
\begin{equation}
\epsilon_t^F \leq \eta,
\qquad
\epsilon_t^A \leq \eta,
\end{equation}
where
\begin{equation}
\epsilon_t^F
=
\|F(u_t)-\Phi_{\Delta t}(u_t)\|,
\end{equation}
and
\begin{equation}
\epsilon_t^A
=
\left\|
\mathrm{Dec}\!\left(z_t+\Delta t\,v_\theta(z_t,b)\right)
-
u_{t+\Delta t}
\right\|,
z_t=\mathrm{Enc}(u_t).
\end{equation}
This assumption does not favor AFNO in terms of single-step prediction accuracy.

For AFNO, we decompose the one-step error budget into a latent dynamical component and a reconstruction component:
\begin{equation}
\eta_{\mathrm{dyn}}=w\eta,
\qquad
\eta_{\mathrm{rec}}=(1-w)\eta,
\qquad
0<w<1.
\end{equation}
Here, $\eta_{\mathrm{dyn}}$ represents the error injected into latent evolution, while $\eta_{\mathrm{rec}}$ represents the encoder--decoder reconstruction contribution.
Unlike autoregressive physical-space rollout, the reconstruction contribution is not repeatedly fed back into the latent dynamics at intermediate rollout steps.

\noindent\textbf{\textit{Assumption 5 (Bounded average latent propagation gain).}}
Let $D_i^A$ denote the physical-space error contribution at the $i$-th rollout step induced by propagated latent dynamical errors after decoding.
We assume that the average propagation gain of AFNO latent evolution is bounded independently of the rollout length $K$; that is, there exists a constant $\kappa_A=O(1)$ such that
\begin{equation}
\sum_{i=1}^{K}D_i^A
\leq
\kappa_A K\eta_{\mathrm{dyn}}
=
\kappa_A K w\eta .
\end{equation}
This assumption does not require every rollout step to be contractive.
It only requires that the accumulated propagated latent dynamical error grows at most linearly with the rollout length.
Such a condition is consistent with the regularity of the learned latent vector field, the local consistency enforced by $\mathcal{L}_{\mathrm{flow}}$, and the multi-step constraint introduced by $\mathcal{L}_{\mathrm{roll}}$.

\textit{Proposition 3 (Structural comparison under simplified accumulation conditions).}
Under Assumptions 4 and 5, consider a neutral accumulation regime for autoregressive physical-space rollout, where the one-step prediction error is repeatedly injected and is not canceled by additional contraction. 
In this regime, the $i$-th step state error of the FNO-type autoregressive predictor can be characterized as
\begin{equation}
e_i^F \sim i\eta .
\end{equation}
Define the accumulated rollout errors over $K$ prediction steps as
\begin{equation}
E_K^F=\sum_{i=1}^{K} e_i^F,\qquad
E_K^A=\sum_{i=1}^{K} e_i^A .
\end{equation}
Then the autoregressive physical-space rollout satisfies
\begin{equation}
E_K^F=\Theta(K^2\eta),
\end{equation}
whereas AFNO satisfies
\begin{equation}
E_K^A \leq K[(1-w)+\kappa_A w]\eta = O(K\eta).
\end{equation}
Accordingly, the accumulated rollout error ratio is lower bounded as
\begin{equation}
\frac{E_K^F}{E_K^A}
\gtrsim
\frac{K}{2[(1-w)+\kappa_A w]}
=
\Omega(K).
\end{equation}
This comparison suggests that, under comparable one-step consistency and bounded average latent propagation gain, AFNO admits a more favorable accumulated error scaling than autoregressive physical-space rollout.

\noindent\textit{Proof.}
For the FNO-type autoregressive predictor, the complete one-step physical-space error is injected at every rollout step.
In the neutral accumulation regime, this gives
\begin{equation}
e_i^F\sim i\eta .
\end{equation}
Therefore,
\begin{equation}
\mathcal{E}_K^F
=
\sum_{i=1}^{K}e_i^F
\sim
\sum_{i=1}^{K}i\eta
=
\frac{K(K+1)}{2}\eta
=
\Theta(K^2\eta).
\end{equation}

For AFNO, the error at the $i$-th output state consists of two parts.
The first part is the reconstruction contribution, whose scale is bounded by $\eta_{\mathrm{rec}}=(1-w)\eta$.
The second part is the propagated latent dynamical error, denoted by $D_i^A$.
Thus,
\begin{equation}
e_i^A
\lesssim
(1-w)\eta+D_i^A .
\end{equation}
Summing over $i=1,\ldots,K$ yields
\begin{equation}
\mathcal{E}_K^A
\lesssim
K(1-w)\eta
+
\sum_{i=1}^{K}D_i^A .
\end{equation}
By Assumption 5,
\begin{equation}
\sum_{i=1}^{K}D_i^A
\leq
\kappa_A K w\eta .
\end{equation}
Therefore,
\begin{equation}
\mathcal{E}_K^A
\lesssim
K(1-w)\eta
+
\kappa_A K w\eta
=
K\left[(1-w)+\kappa_A w\right]\eta .
\end{equation}
Hence,
\begin{equation}
\mathcal{E}_K^A=O(K\eta).
\end{equation}
Combining the FNO and AFNO estimates gives
\begin{equation}
\frac{\mathcal{E}_K^F}{\mathcal{E}_K^A}
\gtrsim
\frac{\frac{K(K+1)}{2}\eta}
{K\left[(1-w)+\kappa_A w\right]\eta}
\sim
\frac{K}{2\left[(1-w)+\kappa_A w\right]},
\end{equation}
which implies
\begin{equation}
\frac{\mathcal{E}_K^F}{\mathcal{E}_K^A}
=
\Omega(K).
\end{equation}
This completes the proof.

\noindent\textbf{\textit{Remark.}}
This result is intended as a structural comparison rather than a universal guarantee of linear error growth.
The advantage of AFNO does not rely on smaller one-step errors, but on avoiding recurrent injection of reconstruction errors during latent-space evolution.
With bounded average latent propagation gain $\kappa_A$, this structure explains the milder empirical rollout error growth.

\section{Additional Visualizations}
\label{sec:5.9}

This section provides additional qualitative visualizations to complement the quantitative results and the main visual comparisons reported in the paper. 
The visualizations cover both one-dimensional and two-dimensional time-dependent PDEs, including 1D Burgers, 1D Allen-Cahn, 2D Navier-Stokes,  and 2D complex Ginzburg-Landau equations. 
For each benchmark, we present time-evolving predicted solutions and, when applicable, the corresponding error maps to examine how prediction errors develop over long-horizon rollout.

Specifically, Figs.~\ref{fig:9} and~\ref{fig:10} provide analogous comparisons for the 1D Burgers and 1D-Allen-Cahn equations, respectively. 
Figs.~\ref{fig:12} and~\ref{fig:13} present the predicted solutions and error maps for the 2D Navier-Stokes equations, while Figs.~\ref{fig:14} and~\ref{fig:15} report the corresponding results for the 2D complex Ginzburg-Landau equation. 
Overall, these results show that AFNO produces predictions that remain visually closer to the reference solutions and exhibit more stable error patterns over extended temporal horizons.

\begin{figure*}[h]
  \centering
  \includegraphics[width=\linewidth]{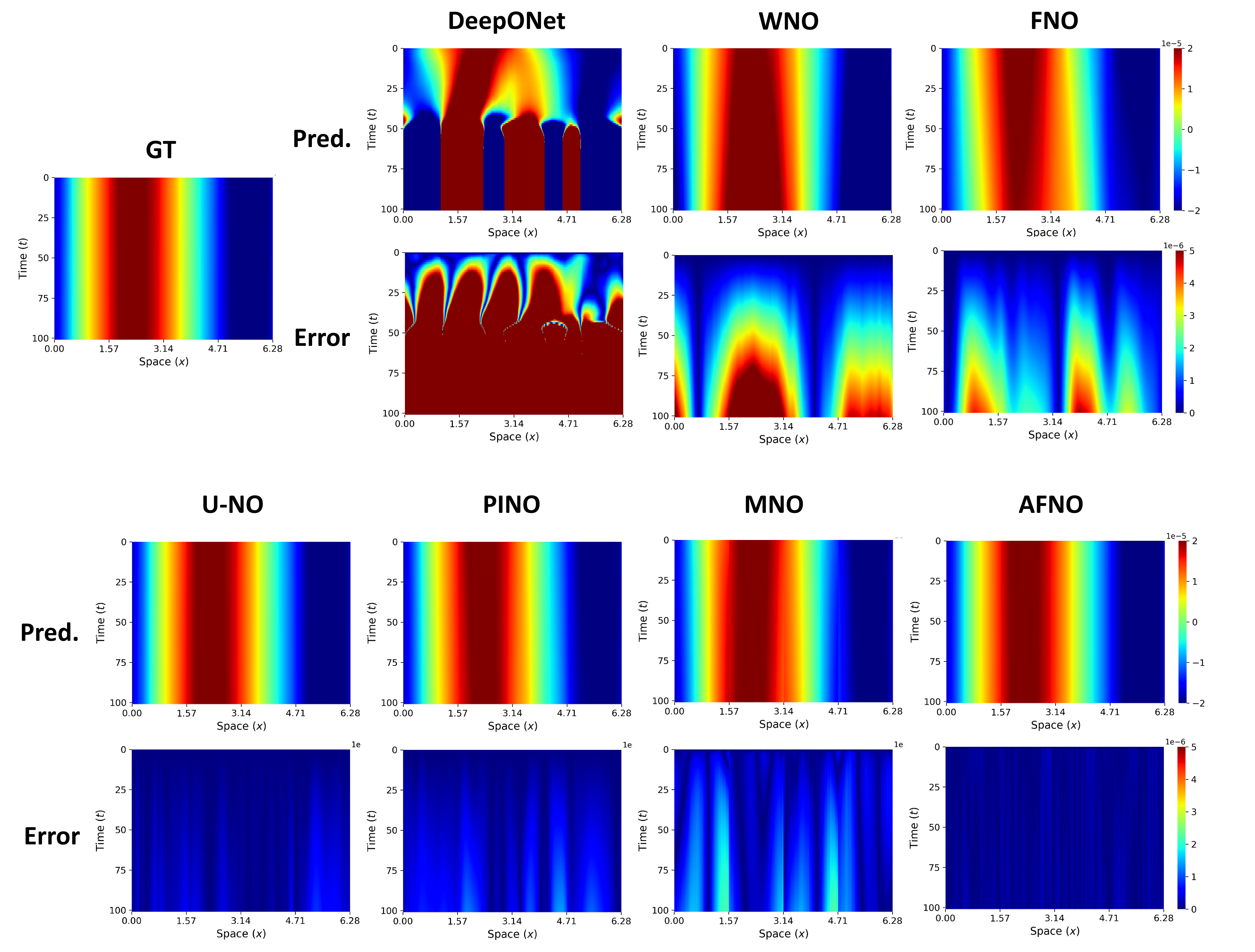}
   \caption{Comparison of time-evolving predicted solutions and corresponding error maps for the 1D Burgers equation.}
  \label{fig:9}
\end{figure*}
\begin{figure*}[h]
  \centering
  \includegraphics[width=\linewidth]{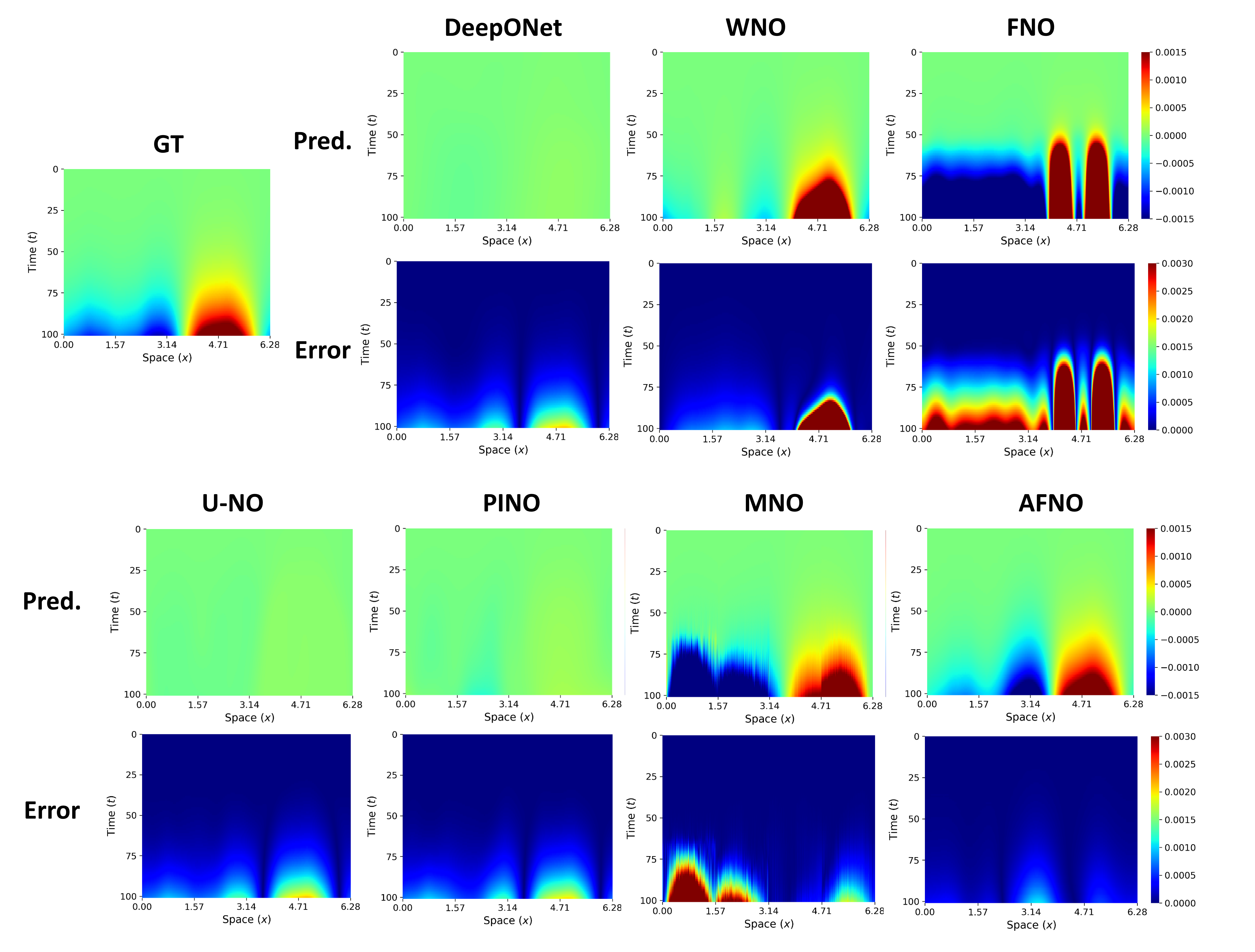}
   \caption{Comparison of time-evolving predicted solutions and corresponding error maps for the 1D Allen-Cahn equation.}
  \label{fig:10}
\end{figure*}

\begin{figure*}[h]
  \centering
  \includegraphics[width=\linewidth]{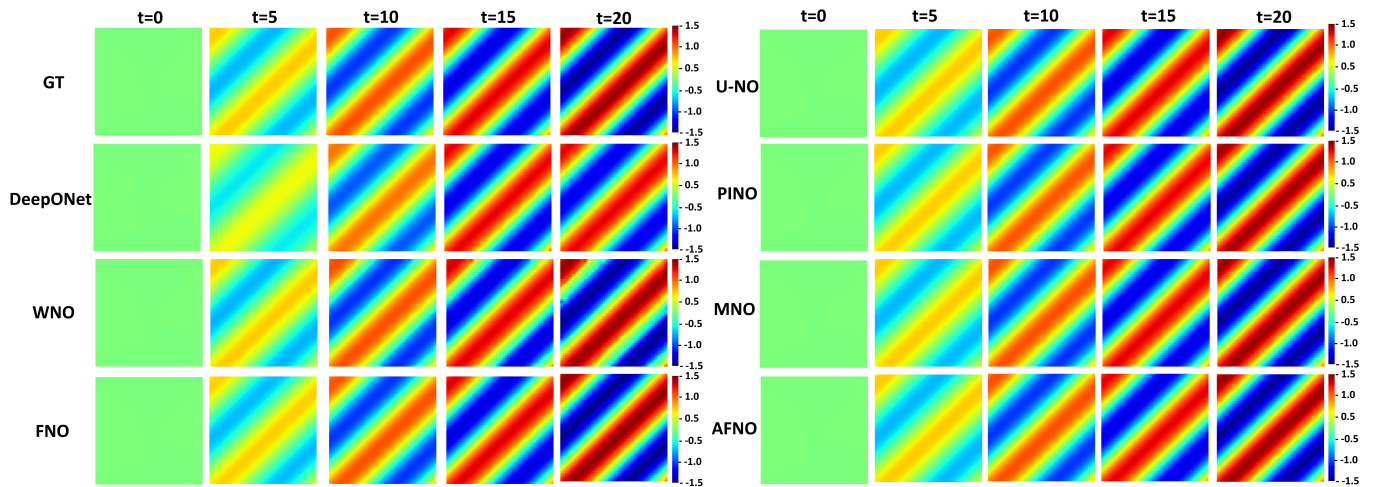}
   \caption{Qualitative comparison of predicted solutions over time for the 2D Navier-Stokes equations.}
  \label{fig:12}
\end{figure*}
\begin{figure*}[h]
  \centering
  \includegraphics[width=\linewidth]{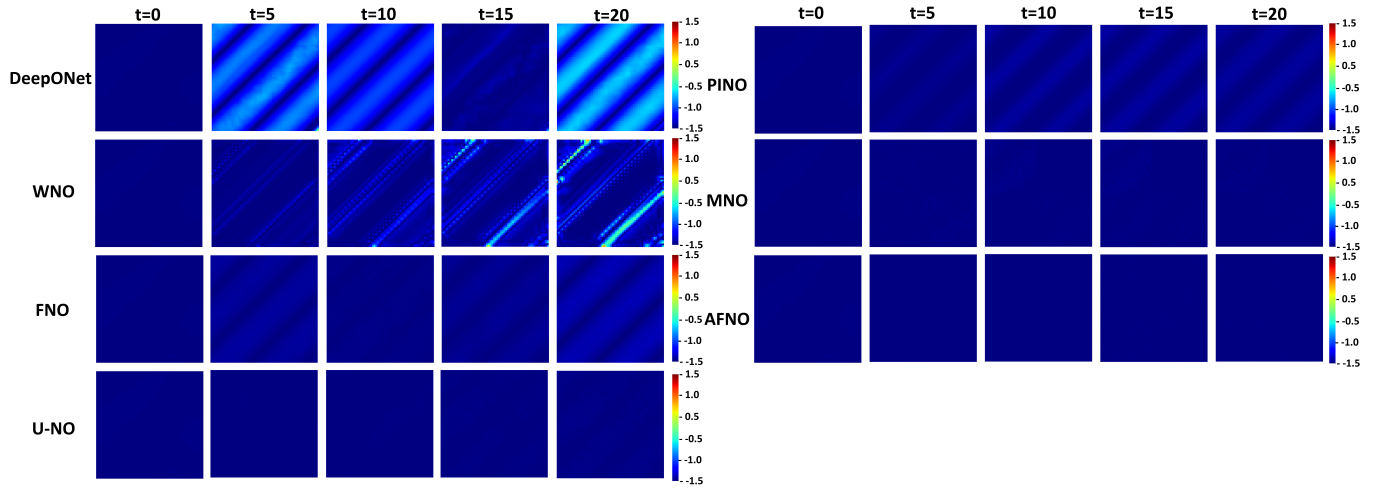}
   \caption{Qualitative comparison of error maps over time for the 2D Navier-Stokes equations.}
  \label{fig:13}
\end{figure*}
\begin{figure*}[h]
  \centering
  \includegraphics[width=\linewidth]{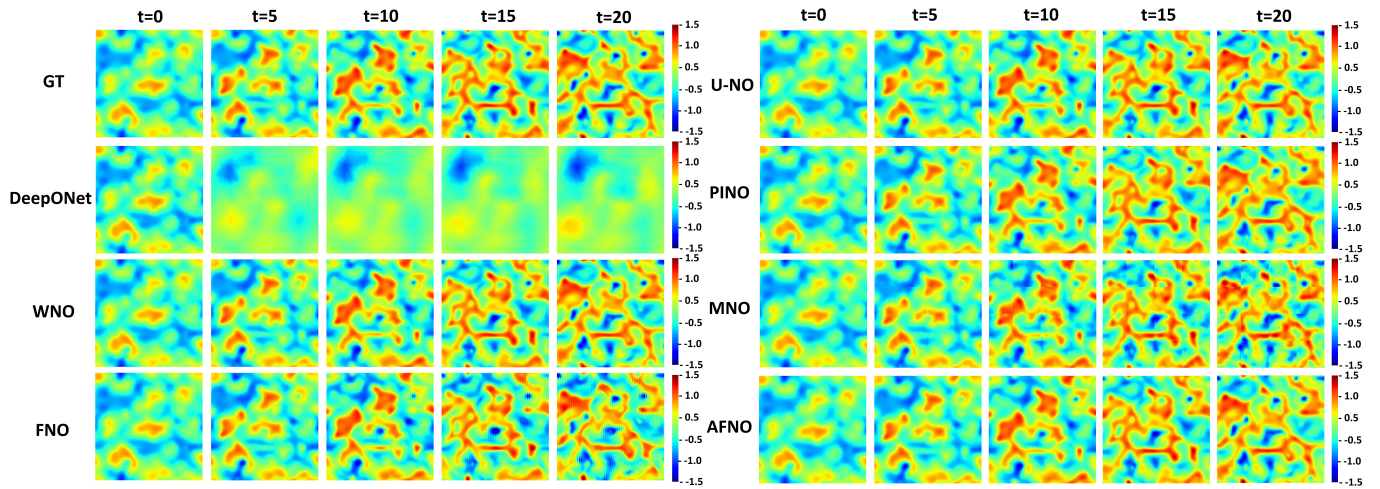}
   \caption{Qualitative comparison of predicted solutions over time for the 2D complex Ginzburg-Landau equation.}
  \label{fig:14}
\end{figure*}
\begin{figure*}[h]
  \centering
  \includegraphics[width=\linewidth]{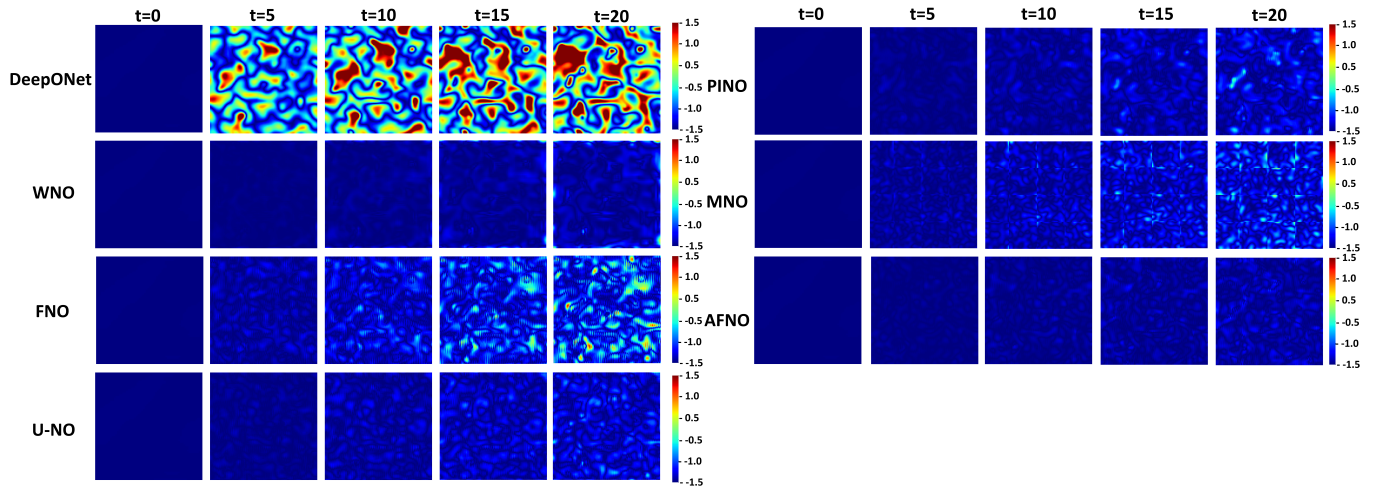}
   \caption{Qualitative comparison of error maps over time for the 2D complex Ginzburg-Landau equation.}
  \label{fig:15}
\end{figure*}

}

\end{document}